\documentclass[runningheads]{llncs}

\usepackage{eccv}

\usepackage{eccvabbrv}

\usepackage{graphicx}
\usepackage{booktabs}
\usepackage{makecell}
\usepackage{microtype}   % helps fit more text

\usepackage[accsupp]{axessibility}  % Improves PDF readability for those with disabilities.

\usepackage[hidelinks]{hyperref}

\usepackage{orcidlink}

\definecolor{commentblue}{rgb}{0.0196,    0.4863,    0.8784}

\definecolor{todored}{rgb}{0.8,    0.263,    0.184}

\definecolor{lightblue}{rgb}{.8,.95,1}
\newcommand{\psnr}{\scalebox{0.8}{PSNR$\uparrow$}}
\newcommand{\FPS}{\scalebox{0.8}{FPS$\uparrow$}}
\newcommand{\mIoU}{\scalebox{0.8}{mIOU$\uparrow$}}
\newcommand{\mAcc}{\scalebox{0.8}{mACC$\uparrow$}}

\newcommand{\dssim}{\scalebox{0.8}{DSSIM$\downarrow$}}
\newcommand{\first}[1]{{\colorbox{red!40}{#1}}}
\newcommand{\second}[1]{{\colorbox{orange!50}{#1}}}
\newcommand{\third}[1]{{\colorbox{yellow!50}{#1}}}

\begin{document}

\title{Multi4D: High-Fidelity Dynamic Gaussian Splatting via Multi-Level Competitive Allocation} 

\titlerunning{Multi4D}

\author{Rui Wang\orcidlink{0009-0008-4761-0670}\thanks{Corresponding author.}  \and Quentin Lohmeyer\orcidlink{0000-0003-3802-5329} \and  Siyu Tang\orcidlink{0000-0002-1015-4770}  \and Mirko Meboldt\orcidlink{0000-0001-5828-5406} \\
}
% \author{Rui Wang \orcidlink{0000-1111-2222-3333} \and
% Quentin Lohmeyer\inst{2,3}\orcidlink{1111-2222-3333-4444} \and
% Third Author\inst{3}\orcidlink{2222--3333-4444-5555}}

% % TODO FINAL: Replace with an abbreviated list of authors.
\authorrunning{R.~Wang et al.}

% % TODO FINAL: Replace with your institution list.

\institute{ ETH Z\"urich, Zurich ZH 8092, Switzerland \\
\email{ \{ruiwang46, qlohmeye, meboldt\}@ethz.ch} \quad\quad \email{siyu.tang@inf.ethz.ch} \\
\url{https://batfacewayne.github.io/Multi4D.io/}
}

\maketitle

\begin{abstract}
Dynamic 3D Gaussian splatting faces a fundamental tension between motion consistency and visual fidelity. Deformation-based approaches preserve temporal correspondence but suffer from motion over-factorization, oversmoothing high-frequency dynamics. In contrast, 4D-primitive methods capture fine visual details yet incur temporal over-parameterization, breaking object identity and leading to severe storage overhead.
To resolve this, we introduce Multi4D, a framework for high-fidelity dynamic Gaussian Splatting based on multi-level competitive allocation. Instead of a monolithic representation, we distribute modeling capacity across three structured levels: static structure, persistent dynamic geometry, and transient appearance primitives. Through shared rasterization and residual-driven optimization, these levels dynamically compete to explain photometric error, enabling adaptive specialization without pre-assigned decomposition.
This allocation preserves long-term motion consistency while capturing fine dynamic detail, achieving state-of-the-art rendering quality and real-time performance with significantly fewer dynamic primitives. Furthermore, because our representation explicitly tracks compact persistent Gaussians over time, semantic features can be embedded afterward, enabling Multi4D to achieve state-of-the-art 4D segmentation accuracy with an order-of-magnitude speedup.

\end{abstract}

\section{Introduction}
\label{sec:intro}
\begin{figure*}[t]
\includegraphics[width=\linewidth]{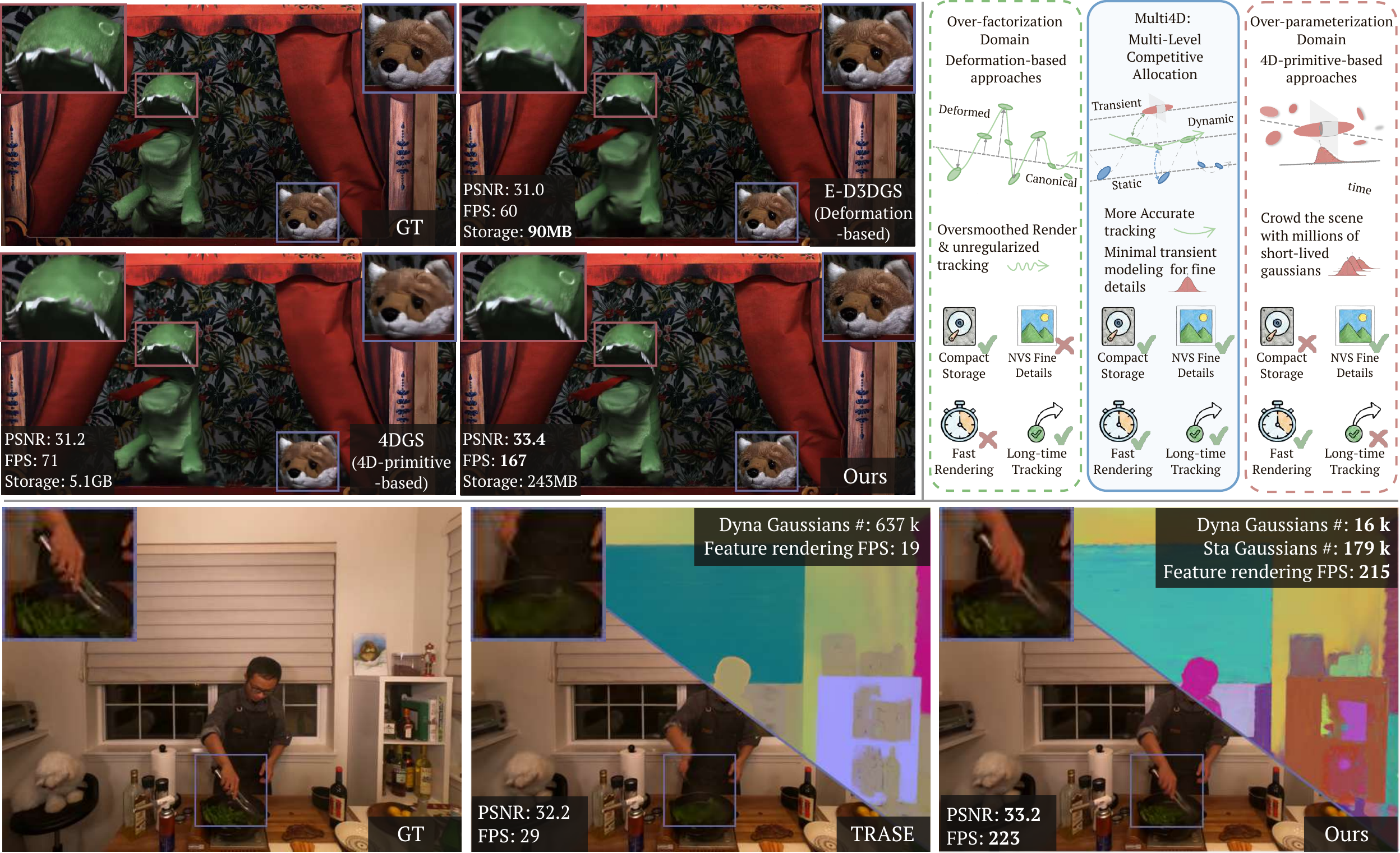}
\captionof{figure}{\textbf{Multi4D} enables (1) high-quality, efficient dynamic scene reconstruction via competitive multi-level specialization, and (2) compact, high-accuracy 4D segmentation with fast inference.}
\label{fig:teaser}
\end{figure*}
3D Gaussian Splatting (3DGS)~\cite{kerbl20233d} has revolutionized novel view synthesis by combining explicit 3D primitives with real-time differentiable rasterization. However, extending this representation to \textit{dynamic} scenes exposes a fundamental conflict between physical plausibility (tracking and correspondence) and visual fidelity (rendering quality).
Current dynamic 3DGS frameworks generally fall into two opposing domains. The first, comprising \textit{deformation-based} approaches, maintains a fixed set of canonical Gaussians and warps them over time via neural deformation networks~\cite{yang2024deformable}, explicit trajectory modeling~\cite{luiten2024dynamic,huang2024sc}, or feature grids~\cite{wu20244d,yang2024deformable}. By enforcing a strict temporal correspondence, these methods inherently preserve Gaussian identity, making them suited for downstream tasks like semantic embedding and tracking~\cite{li2024trase,ji2024segment}. However, this formulation often leads to severe \emph{motion over-factorization}, where the deformation field groups nearby motion and over-smooths high-frequency dynamics.
While recent works attempt to alleviate this over-smoothing via coarse-to-fine deformation~\cite{kwak2025modec}, frequency-aware fields~\cite{miaofrequency}, or spline-based trajectories~\cite{yoon2025splinegs}, they remain fundamentally constrained by two major drawbacks: Because the representation is dominated by motion modeling, these methods often interpret complex appearance changes (e.g., specularities or lighting shifts) as physical motion, causing spurious geometric warping to minimize photometric error. Moreover, they face a non-trivial computational bottleneck: querying the deformation network for \textit{every} primitive at \textit{every} timestamp incurs an overhead that scales strictly with the total number of Gaussians. Consequently, increasing the primitive count to capture fine dynamic details
  directly compromises real-time rendering capacity and memory efficiency~\cite{wang2025degauss,TuYing2025SpeeDe3DGS}.

Conversely, the second domain comprises \textit{4D-primitive}~\cite{yang2023real,duan20244d}, which conceptualize scene dynamics as 4D spatiotemporal Gaussian hyper-cylinders. At any given timestamp, these primitives are sliced along the temporal axis to yield 3D Gaussians with time-varying opacities and linear local position shifts. By modulating temporal existence, these methods seamlessly capture complex appearance changes and transient geometry. Furthermore, this time-dependent formulation enables temporal pre-filtering to render only active Gaussians, accelerating rendering speed. However, while effectively modeling the dynamic scene with fine details by piece-wise linear approximation, these methods lead to severe \emph{temporal over-parameterization}. They are prone to modeling physical motion as transient appearance changes, where the optimization tends to exploit temporal scaling to minimize photometric error, hallucinating millions of primitives that exist for extremely short lifespans rather than capturing physical movement, leading to broken geometry in the fast motion region. Most recent methods~\cite{lee2025optimized,zhang2025mega,yuan20251000+} attempt to reduce the number of Gaussian primitives via temporal score pruning or sensitivity analysis, primarily for static regions to alleviate the storage overhead at the cost of slightly reducing rendering quality. However, these approaches are generally limited by the local piece-wise motion approximation assumption and cannot further efficiently simplify dynamic regions. Moreover, lacking holistic modeling and canonical geometric constraints limits these methods to only dense camera input, and generalizes poorly to sparse camera inputs and monocular settings, compared to deformation-based approaches.

We therefore challenge the monolithic assumption that one representation must explain both physical kinematics and transient appearance. Instead, we formulate dynamic reconstruction as a \emph{competitive multi-level optimization problem}, in which models with distinct inductive biases dynamically compete to explain photometric residuals. By explicitly separating statics from dynamics, and geometric deformation from transient appearance, we enable residual-driven allocation across levels under a unified differentiable renderer.
To this end, we introduce \textbf{Multi4D}, a competitive multi-level framework that jointly optimizes three functionally specialized Gaussian subsets: 
(1) \textbf{Static Gaussians} ($\mathcal{G}_s$), which provide a time-invariant structural backbone; 
(2) \textbf{Persistent Dynamic Gaussians} ($\mathcal{G}_d$), deformable primitives governed by a holistic deformation module to maintain long-term identity and trackability; 
and (3) \textbf{Transient Gaussians} ($\mathcal{G}_t$), short-lived 4D primitives dedicated exclusively to modeling high-frequency appearance residuals. A high-level overview of our proposed method is shown in~\cref{fig:teaser}.

Through shared rasterization, gradients are coupled across subsets: once stable geometry explains a region, residual-driven allocation suppresses redundant modeling in other subsets. This structural decoupling prevents deformation fields from absorbing transient photometric noise while allowing high-capacity 4D primitives to model irreducible appearance variation. As a result, Multi4D preserves long-term motion consistency and fine-grained dynamic detail simultaneously.
By explicitly separating structural persistence from transient modeling, dynamic novel view synthesis benefits from coherent geometry with significantly fewer modeled primitives, yielding higher fidelity, faster rendering, and improved generalization under sparse or monocular supervision. For 4D segmentation, we restrict optimization to the persistent subset $\mathcal{G}_s \cup \mathcal{G}_d$, which maintains stable temporal identity while discarding transient appearance noise. This produces compact feature rendering, stronger motion consistency, and state-of-the-art tracking accuracy with an order-of-magnitude speedup.
\noindent\textbf{Our contributions are summarized as follows:}
\begin{itemize}
\item We propose \textbf{Multi4D}, a multi-level Gaussian decomposition framework that separates static structure, persistent dynamic geometry, and transient appearance modeling, reconciling motion consistency with photometric fidelity.

\item We introduce a bottom-up, self-regularized training strategy with velocity-aware lifting and mask-aware pruning, enabling structured specialization and compact representations.

\item We demonstrate that Multi4D achieves state-of-the-art dynamic novel view synthesis, delivering superior rendering fidelity with significantly fewer dynamic primitives and improved runtime performance.

\item We further show that Multi4D naturally supports state-of-the-art 4D segmentation with significantly faster inference.
\end{itemize}
\section{Related Work}
\label{sec:related_work}
\textbf{Deformation-based dynamic Gaussian methods} track a canonical set of 3D Gaussians using time-conditioned deformation modules, including explicit trajectories~\cite{luiten2024dynamic,huang2024sc}, neural networks~\cite{yang2024deformable}, and grids~\cite{cao2023hexplane,wu20244d}. Extensions introduce finer deformation models such as multi-grid refinement, spline trajectories, and frequency-aware networks~\cite{xu2024grid4d,fan2025spectromotion,yoon2025splinegs,miaofrequency}. However, per-frame querying deformation fields for every primitive incurs significant computational overhead and limits modeling capacity. Efficiency-oriented variants decouple static and dynamic~\cite{wu2025swift4d,wang2025degauss}, yet complex time-varying appearance modeling remains challenging.
% \textbf{Deformation-based dynamic gaussian methods} model the dynamic scene by tracking a canonical set of 3d gaussians and per-frame deformation modules, including explicit trajectory modeling~\cite{luiten2024dynamic, huang2024sc}, time-conditioned neural network~\cite{yang2024deformable} or spatial-temporal encoded grids~\cite{cao2023hexplane, wu20244d}. While providing a strong baseline to versatile inputs, these methods fail to capture fine-grained motion and appearance changes. To alleviate these artifacts, methods as~\cite{xu2024grid4d,fan2025spectromotion} employ a multi-grid architecture for finer spatial and temporal feature aggregation.Spectromotion~\cite{fan2025spectromotion} employs residual correction for specular objects modeling. Spline-GS~\cite{yoon2025splinegs} leverages spline-based trajectories coupled with a hash table and MLP to model gaussian deformation. Most recently, FAGS~\cite{miaofrequency} leverages a Fourier-Deformation Network to capture high-frequency local deformations. However, querying a neural field or feature grid for \emph{every} primitive at \emph{every} frame creates a severe computational bottleneck. Recent methods like Swift4D~\cite{wu2025swift4d} and DeGauss~\cite{wang2025degauss} attempt to alleviate this by explicitly decoupling dynamic from static scene modeling to improve efficiency, yet they still struggle with complex topological breaks.

\textbf{4D-primitive-based methods} represent dynamic scenes using spatiotemporal Gaussians sliced at each timestamp~\cite{yang2023real,duan20244d}. Although capable of high-fidelity rendering, this formulation often leads to temporal over-parameterization, spawning large numbers of short-lived primitives. Extensions improve compactness and temporal modeling via feature-based rendering, explicit velocity parameters, or static–dynamic factorization~\cite{li2024spacetime,wang2025freetimegs,lee2024fully,oh2025hybrid}. Temporal pruning and sensitivity analysis further reduce redundancy~\cite{lee2025optimized,zhang2025mega,yuan20251000+}. However, the lack of persistent geometric tracking limits compression in highly dynamic regions and reduces robustness under sparse or monocular supervision.
% \textbf{4D-primitive based approaches} model the scene at any given timestamp with 4D primitives sliced to yield 3D Gaussians with time-varying positions and opacity~\cite{yang2023real,duan20244d}. 
% While enabling high-fidelity rendering, the optimization tends to exploit this slicing mechanism to spawn millions of short-lived primitives, leading to \emph{temporal over-parameterization}. STG~\cite{li2024spacetime} further uses splatted feature rendering to enhance model compactness. FreeTimeGS~\cite{wang2025freetimegs} attempts to improve temporal continuity by equipping 4D primitives with explicit velocity parameters; however, this approach remains bottle-necked by its reliance on pre-computed point tracking and dense, per-frame 3D initialization. Recognizing the severe memory bloat of these 4d-primitve approaches, recent hybrid approaches~\cite{lee2024fully, oh2025hybrid} explicitly factorize static 3D Gaussians from the dynamic scene to achieve more compact representations. several works~\cite{lee2025optimized,zhang2025mega,yuan20251000+} employ temporal score pruning or sensitivity analysis to aggressively cull redundant primitives. However, as the 4d-primitive approaches lacks long-term motion tracking capabilities, these pruning heuristics fail to compress highly dynamic regions without degrading visual quality. Furthermore, without a persistent global geometric prior, these methods continue to struggle significantly on sparse or monocular inputs.

\textbf{Semantic Embedding with Gaussian Splatting} combines Gaussian representations with foundation models~\cite{kirillov2023segment,caron2021emerging} for 3D scene understanding. Extensions to dynamic settings build upon deformation-based frameworks to maintain temporal identity~\cite{ji2024segment,li20254d,labe2024dgd,li2024trase}. However, these approaches inherit the computational overhead of deformation architectures, as high-dimensional semantic features must be evaluated and rendered for every primitive at each frame.

These limitations motivate representations that preserve persistent geometry, maintain sparse parameterization, and enable high-fidelity dynamic rendering.
% \textbf{Semantic Embedding with Gaussian Splatting} has been developed for 3D scene understanding in combination with 2D foundation models like SAM~\cite{kirillov2023segment} and DINO~\cite{caron2021emerging}. While popular for static scenes, only a few methods extend this to dynamic environments. SA4D~\cite{ji2024segment} and 4D-LangSplat~\cite{li20254d} extend deformation-based approaches~\cite{yang2024deformable,wu20244d} to devise temporal identity feature fields. DGD~\cite{labe2024dgd} introduces high-dimensional feature rendering to Deformable-3DGS, and most recently, TRASE~\cite{li2024trase} adopts a similar approach augmented with KNN feature smoothing. However, these methods inherently suffer from the severe computational bottlenecks of their underlying deformation architectures. This slow inference is drastically exacerbated by the requirement to evaluate and render high-dimensional semantic features for \emph{every} primitive at \emph{every} frame. 
\section{Method}
\label{sec:method}
\begin{figure*}[t]
  \centering
  \includegraphics[width=\linewidth]{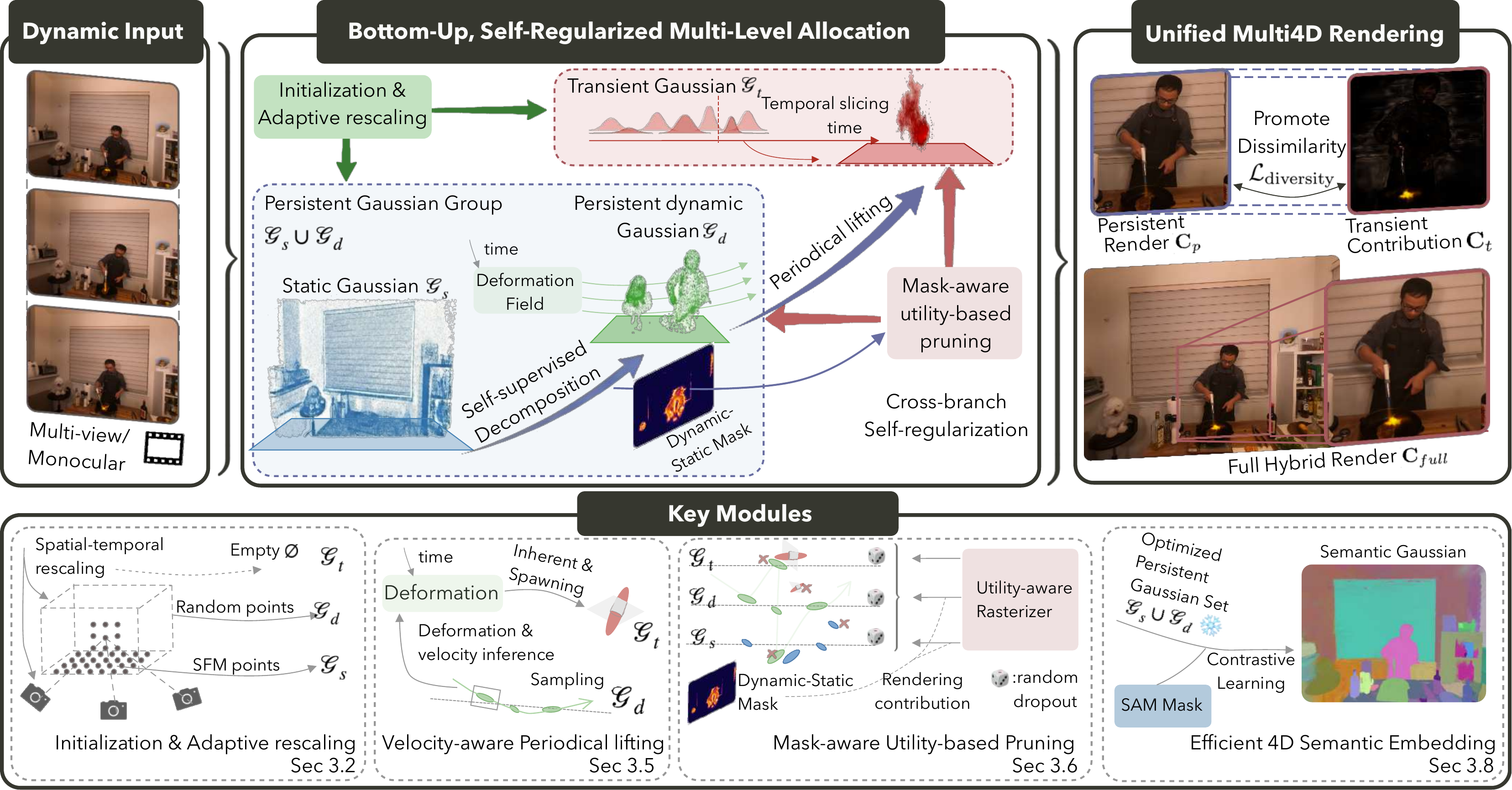}

\caption{\textbf{Overview of the Multi4D pipeline.} 
We employ a bottom-up training scheme that enables competitive allocation across multi-level Gaussian subsets through cross-set self-supervision. 
After optimization, the persistent subset $\mathcal{G}_p$ is frozen and used for efficient downstream 4D segmentation.
}
  \label{fig:pipeline}
\end{figure*}
We formulate dynamic reconstruction over $\mathcal{S} = \{I_t\}_{t=1}^T$ using a bottom-up multi-level competitive allocation framework. Instead of prescribing fixed roles, modeling responsibility emerges through optimization among three specialized Gaussian subsets that compete under a shared photometric objective.

We represent the scene as $\mathcal{G} = \mathcal{G}_s \cup \mathcal{G}_d \cup \mathcal{G}_t$:
\textbf{Static ($\mathcal{G}_s$)}: time-invariant 3D Gaussians anchoring stable structure.
\textbf{Persistent Dynamic ($\mathcal{G}_d$)}: canonical Gaussians driven by a geometric-only deformation field $\Phi_g$  based on HexPlane~\cite{cao2023hexplane,wu20244d} that predicts rigid motion:
$ (\boldsymbol{\mu}_t, \mathbf{r}_t) = (\boldsymbol{\mu}, \mathbf{r}) + \Phi_g(\boldsymbol{\mu}, t) $,
preserving temporal identity without appearance drift.
\textbf{Transient Dynamic ($\mathcal{G}_t$)}: 4D spatiotemporal Gaussians modeling high-frequency appearance changes and transient geometry.

An overview of the pipeline is shown in~\cref{fig:pipeline}. All subsets are jointly rendered within a unified differentiable rasterizer, where shared transmittance couples their gradients and naturally induces competition across subsets. Once one subset explains a region, residual-driven densification in the remaining subsets is suppressed. The following sections detail the resulting bottom-up multi-level optimization strategy.
\subsection{Preliminaries: 3D \& 4D Gaussian Splatting}
\label{sec:preliminaries}
\textbf{3D Gaussian Splatting}~\cite{kerbl20233d} provides an explicit representation of a static 3D scene using a set of anisotropic Gaussian primitives $\mathcal{G}$. Each primitive $g \in \mathcal{G}$ is defined by a mean vector \(\boldsymbol{\mu}\) and a covariance matrix \(\mathbf{\Sigma}\), formulated as:
\begin{equation}
    \label{eqn:gaussian_geometry}
    g(\mathbf{x}) = \exp \Bigl( -\frac{1}{2} \left( \mathbf{x} - \boldsymbol{\mu} \right)^T \mathbf{\Sigma}^{-1} \left( \mathbf{x} - \boldsymbol{\mu} \right) \Bigr), \quad \text{s.t.} \quad \mathbf{\Sigma} = \mathbf{R}\mathbf{S}\mathbf{S}^T \mathbf{R}^T,
\end{equation}
where \(\mathbf{R}\) is the rotation matrix (derived from a quaternion $\mathbf{r}$) and \(\mathbf{S} = \text{diag}(\mathbf{s})\) is the scale matrix, respectively. Differentiable splatting~\cite{yifan2019differentiable} is used to render these Gaussians onto the image plane. The final color \(\mathbf{C}\) at pixel \(\mathbf{u}\) is computed by blending the contributions of all $N$ depth-sorted Gaussians:
\begin{equation}
    \label{eqn:gaussian_splatting}
    \mathbf{C}(\mathbf{u})
    = \sum_{i=1}^{N} \mathbf{c}_i(\mathbf{d}) \,\sigma_i \,\mathcal{P}_i(g_i,\mathbf{u})
      \prod_{j=1}^{i-1}(1 - \sigma_j \,\mathcal{P}_j(g_j,\mathbf{u})),
\end{equation}
with \(\mathbf{c}_i(\mathbf{d})\in\mathbb{R}^k\) denoting spherical harmonic (SH) coefficients evaluated at the view direction $\mathbf{d}$, \(\sigma_i\) denoting the opacity, and $\mathcal{P}_i$ denoting the 2D projection of each Gaussian $g_i$.

\textbf{4D Gaussian Splatting.} To capture temporal appearances without explicit deformation fields, native 4D approaches~\cite{yang2023real} extend Gaussian primitives into spacetime using 4D ellipsoids defined by \(\boldsymbol{\mu}_{4D} = (\boldsymbol{\mu}_{xyz}, \mu_t)\in \mathbb{R}^4\) and a 4D covariance matrix \(\mathbf{\Sigma}_{4D}\). At a given timestamp $t$, the time-conditioned 3D Gaussian geometry is analytically derived via slicing:
\begin{equation}
\label{equ:marginal}
\begin{split}
    \boldsymbol{\mu}_{xyz|t} &= \boldsymbol{\mu}_{1:3} + \mathbf{\Sigma}_{1:3,4} \mathbf{\Sigma}_{4,4}^{-1} (t - \mu_t), \\
    \mathbf{\Sigma}_{xyz|t} &= \mathbf{\Sigma}_{1:3,1:3} - \mathbf{\Sigma}_{1:3,4} \mathbf{\Sigma}_{4,4}^{-1} \mathbf{\Sigma}_{4,1:3}.
\end{split}
\end{equation}
Moreover, the view-dependent color is extended to 4D spherical harmonic (4DSH)~\cite{yang2023real}, capturing time-varying appearance with $\mathbf{c}_{4D}(\mathbf{d}, t - \mu_t)$.
\subsection{Initialization and Adaptive Rescaling}
\label{sec:initialization}

\textbf{Initialization via Inverse Expressiveness.}
To enable bottom-up specialization during optimization, we impose an inductive bias hierarchy by initializing each Gaussian subset inversely proportional to its expressive capacity. The constrained static subset $\mathcal{G}_s$ is densely initialized using COLMAP~\cite{schoenberger2016sfm} points to capture stable scene structure, preventing static elements from being incorrectly modeled as dynamic motion during optimization. The persistent dynamic subset $\mathcal{G}_d$ is initialized with sparse, randomly generated points, allowing gradual dynamic geometric refinement under self-supervision. The highly expressive transient subset $\mathcal{G}_t$ is initialized empty and instantiated only through periodic lifting from $\mathcal{G}_d$ in later training stages, preventing overfitting to reconstruction noise.

\textbf{Adaptive Spatial-Temporal Rescaling.} 
Jointly optimizing these three subsets introduces numerical challenges due to inconsistent scene scales. Uncalibrated spatial and temporal extents can lead to gradient imbalance and ill-conditioned 4D covariance matrices, causing severe instabilities (e.g., singularities when inverting $\mathbf{\Sigma}_{4,4}$ in Eq.~\ref{equ:marginal}). To stabilize optimization, we linearly normalize the spatial and temporal domains prior to training using scaling factors derived from the camera distribution and the total video duration.
\subsection{Unified Multi4D Rendering}
\label{sec:hybrid_render}

To ensure consistent depth ordering and coupled optimization, we render all primitives in a single differentiable pass at timestamp $t$. Each primitive $i \in \mathcal{G}$ is projected into an instantaneous 3D state $\Theta_{t,i} = \{\boldsymbol{\mu}_t, \mathbf{\Sigma}_t, \sigma_t, \mathbf{c}_t\}$:
\begin{equation}
    \Theta_{t, i} = 
    \begin{cases} 
        \left( \boldsymbol{\mu}, \mathbf{\Sigma}, \sigma, \mathbf{c}_{3D} \right)  & \text{if } g_i \in \mathcal{G}_s, \\
        \left( \boldsymbol{\mu}_t, \mathbf{\Sigma}_t, \sigma, \mathbf{c}_{3D} \right)  & \text{if } g_i \in \mathcal{G}_d, \\
        \left( \boldsymbol{\mu}_{xyz|t}, \mathbf{\Sigma}_{xyz|t}, \sigma_t, \mathbf{c}_{4D} \right)  & \text{if } g_i \in \mathcal{G}_t.
    \end{cases}
    \label{equ:hybrid_projection}
\end{equation}

Static and persistent geometries are defined explicitly (via $\Phi_g(\boldsymbol{\mu}, t)$ for $\mathcal{G}_d$), while transient geometry ($\mathcal{G}_t$) is obtained through 4D slicing (Eq.~\ref{equ:marginal}). Transient opacity 
$\sigma_t = \sigma \exp\left(-(t - \mu_t)^2 / 2\mathbf{\Sigma}_{4,4}\right)$,
enables discontinuous temporal modeling, while 4DSH ($\mathbf{c}_{4D}$) captures time-varying appearance.

\noindent\textbf{Unified Blending \& Decoupled Outputs.}
Unlike prior binary static–dynamic designs~\cite{lee2024fully,oh2025hybrid,wu2025swift4d}, we base on ~\cite{oh2025hybrid} and further depth-sort all three subsets jointly within a single rasterization. The shared transmittance accumulation couples their gradients and forms the basis of cross-set competition. During the same pass, we extract three outputs without additional sorting:  
(1) \emph{Full Render} ($\mathbf{C}_{full}$), which blends all $\mathcal{G}$ for photometric supervision;  
(2) \emph{Persistent Render} ($\mathbf{C}_p$), which evaluates $\mathcal{G}_s \cup \mathcal{G}_d$ while ignoring $\mathcal{G}_t$ transmittance; and  
(3) \emph{Transient Contribution} ($\mathbf{C}_t$), which accumulates $\mathcal{G}_t$ using the global transmittance of $\mathcal{G}$.
This decoupling isolates transient visibility while preserving correct cross-set occlusion reasoning. We further apply an SSIM-based diversity loss $\mathcal{L}_\text{diversity}$ between $\mathbf{C}_t$ and $\mathbf{C}_p$ to discourage redundant modeling and encourage subset specialization.
\subsection{Self-Supervised Dynamic-Static Decomposition}
\label{sec:decomp}

Following the self-supervised dynamic-static separation strategy introduced in~\cite{wang2025degauss}, we isolate persistent actors from the background without relying on ground-truth tracking labels. We augment $\mathcal{G}_d$ with a base mask logit $m_i$ and predict a time-dependent offset using an MLP $\mathcal{D}_m$:
\[
m'_i(t) = m_i + \mathcal{D}_m(\mathcal{H}(\boldsymbol{\mu}_i, t)).
\]
By substituting SH colors with $m'_i(t)$ during rasterization, we obtain a continuous 2D dynamic mask $\mathbf{M}_d \in [0,1]^{H \times W}$. We then render color images independently from the dynamic ($\mathbf{C}_d$) and static ($\mathbf{C}_s$) subsets, and compute the composite image used for early photometric supervision as
\[
\mathbf{C}_{\text{comp}} = \mathbf{M}_d \odot \mathbf{C}_d + (1 - \mathbf{M}_d) \odot \mathbf{C}_s.
\]

Densely initializing $\mathcal{G}_s$ from COLMAP~\cite{schoenberger2016mvs,schoenberger2016sfm} while initializing $\mathcal{G}_d$ with sparse random points introduces a structural asymmetry that biases the optimization toward assigning static content to $\mathcal{G}_s$. As a result, the dynamic mask $\mathbf{M}_d$ naturally contracts to regions exhibiting temporal variation.

To further enforce this separation, we introduce a spatially-aware opacity penalty $\mathcal{L}_{\alpha}$ that uses the inferred mask $\mathbf{M}_d$ as a spatial template. It penalizes persistent dynamic Gaussians ($\mathcal{G}_d$) projecting into static regions ($\mathbf{M}_d \approx 0$), driving redundant primitives toward zero opacity and enabling their removal during subsequent densification and pruning.
\subsection{Velocity-Aware Periodical Lifting}
\label{sec:adaptive_lift}
Following our bottom-up training strategy, we introduce \textbf{Velocity-Aware Periodic Lifting}, which periodically promotes a small set of \emph{active} persistent Gaussians into the transient subset. Once dynamic--static separation stabilizes, we use the deformed mask logit $m'_i(t)$ (~\cref{sec:decomp}) as an activity score at timestamp $t$. We then sample $K$ candidates from the active set $\{g_i \in \mathcal{G}_d \mid m'_i(t)>\tau\}$ and lift them to $\mathcal{G}_t$. This sampling remains efficient since $\mathcal{G}_d$ is kept sparse throughout optimization, allowing lifting to operate on a compact candidate set.

We lift these candidates into $\mathcal{G}_t$ using \textbf{Momentum Inheritance}. The parent’s instantaneous velocity is estimated via finite differences $\mathbf{v}_i = (\Phi_g(\boldsymbol{\mu}_i, t+\Delta t) - \Phi_g(\boldsymbol{\mu}_i, t)) / \Delta t$,
and used to initialize the new 4D primitive:
\begin{equation}
    \boldsymbol{\mu}_{4D}^{(new)} = [\boldsymbol{\mu}_i(t) + \epsilon, \ t]^T, \quad 
    \mathbf{r}_{4D}^{(new)} \leftarrow \text{Align}(\mathbf{v}_i).
\end{equation}

A small offset $\epsilon$ is added towards the camera center to prevent immediate occlusion by parent gaussians, while $\text{Align}$ orients the 4D principal axis along the spatiotemporal trajectory $[\mathbf{v}_i^T, 1]^T$.
Momentum inheritance provides a strong motion prior for the high-capacity transient subset, mitigating the instability of unconstrained 4D optimization under sparse or monocular supervision. Once promoted, transient primitives autonomously densify to model high-frequency appearance residuals, while mask-aware pruning removes mutually redundant primitives across $\mathcal{G}_d$ and $\mathcal{G}_t$. This controlled handover allows $\mathcal{G}_t$ to absorb non-rigid residual phenomena, while the geometric-only deformation field $\Phi_g$ models rigid motion with stable temporal correspondence.
\subsection{Mask-Aware Utility-Based Pruning}
\label{sec:pruning}

To maintain subset specialization and suppress redundant cross-set modeling, we introduce mask-aware pruning based on each Gaussian’s contribution to the final rendering, overcoming the limitations of opacity-based pruning. For each Gaussian $g_i \in \mathcal{G}$ and view $I$, we define its peak visible contribution:

\begin{equation}
    w_{i,I} = \max_{\mathbf{u} \in I} \bigg( \sigma_i \mathcal{P}_i(g_i, \mathbf{u}) \prod_{j=1}^{i-1} \big(1 - \sigma_j \mathcal{P}_j(g_j, \mathbf{u})\big) \bigg) \cdot M(\mathbf{u}),
\end{equation}

where the gating mask $M(\mathbf{u})$ applies $\mathbf{M}_d$ for $\mathcal{G}_d$, $(1 - \mathbf{M}_d)$ for $\mathcal{G}_s$, and $1$ for $\mathcal{G}_t$. This foreground–background-aware gating ensures that persistent and static primitives contribute only within their assigned regions, preventing cross-set overlap. For the transient subset $\mathcal{G}_t$, the accumulated transmittance spans the entire depth-sorted union, allowing $w_{i,I}$ to incorporate cross-set occlusion reasoning and suppress transient noise hidden behind solid geometry. We aggregate contributions over a window $\mathcal{I}_s$ to obtain a final score:

\begin{equation}
    s_i = \beta \cdot \max_{I \in \mathcal{I}_s} (w_{i,I})  + (1 - \beta) \cdot \frac{1}{|\mathcal{I}_s|}\sum_{I \in \mathcal{I}_s} w_{i, I}.
\end{equation}
Here, $\beta$ balances preserving occasionally important high-contribution
  primitives against removing consistently unused ones.
Primitives with $s_i < \tau_{\mathrm{prune}} $ are removed, eliminating low-utility primitives while preserving essential scene structure. To further discourage view-dependent overfitting, we apply Stochastic Primitive Dropout~\cite{park2025dropgaussian}, randomly disabling primitives during training to encourage cooperative modeling across subsets.
\subsection{Training Strategy and Objectives}
\label{sec:training_strategy}

We optimize Multi4D end-to-end using a two-stage training schedule that first establishes subset specialization and then focuses on rendering refinement. The overall objective is

\begin{equation}
    \mathcal{L}_{total} =
    \mathcal{L}_\text{color}
    + \lambda_{sep}\mathcal{L}_\text{sep}
    + \lambda_{reg}\mathcal{L}_\text{reg}
    + \lambda_{div}\mathcal{L}_\text{diversity}.
\end{equation}
Here, $\mathcal{L}_\text{color}$ denotes photometric supervision, while $\mathcal{L}_\text{sep}$ denotes dynamic–static separation loss, and $\mathcal{L}_\text{reg}$ stands for regularization losses.

\textbf{Phase I (Subset Formation).}
During the early stage, explicit cross-set mechanisms are activated, including dynamic–static separation~\cref{sec:decomp} and velocity-aware periodic lifting~\cref{sec:adaptive_lift}. Together with the unified Multi4D renderer~\cref{sec:hybrid_render} and mask-aware pruning~\cref{sec:pruning}, these mechanisms allow the representation to progressively reorganize across $\mathcal{G}_s$, $\mathcal{G}_d$, and $\mathcal{G}_t$.

\textbf{Phase II (Rendering Refinement).}
In contrast to Phase I, dynamic–static decomposition and lifting are disabled once subset specialization stabilizes. Optimization then focuses on refining geometry and appearance using the unified Multi4D renderer, while mask-aware pruning remains active to remove redundant primitives and maintain a compact representation.

Losses and training details are provided in the supplementary material.
\subsection{Downstream Application: Efficient 4D Segmentation}
\label{sec:semantics}

To demonstrate the downstream utility of our decomposed representation, we perform 4D segmentation by adapting the contrastive feature distillation framework of TRASE~\cite{li2024trase}. We define the semantic scope on the persistent subset $\mathcal{G}_p = \mathcal{G}_s \cup \mathcal{G}_d$, excluding the transient set $\mathcal{G}_t$.

Following appearance convergence, we freeze all geometric parameters and append a learnable semantic feature $\mathbf{f}_i \in \mathbb{R}^{32}$ to each $i \in \mathcal{G}_p$. We render the semantic map $\hat{S}$ via splatting and supervise it using a soft-mined contrastive objective~\cite{li2024trase} against 2D SAM masks $M_{sam}$:
$\mathcal{L}_{sem} = \mathcal{L}_{pos}(\hat{S}, M_{sam}) + \mathcal{L}_{neg}(\hat{S}, M_{sam})$.

Finally, to enforce internal spatial coherence within actors, we apply KNN feature smoothing directly in canonical 3D space:
$\mathbf{f}_i \leftarrow \text{Normalize}\!\left(\frac{1}{|\mathcal{N}_i|}\sum_{j\in\mathcal{N}_i}\mathbf{f}_j\right)$.
Additional details are provided in the supplementary material.
\section{Experiments}
\label{sec:experiments}

\begin{figure*}[t]
  \centering
  \includegraphics[width=0.9\linewidth]{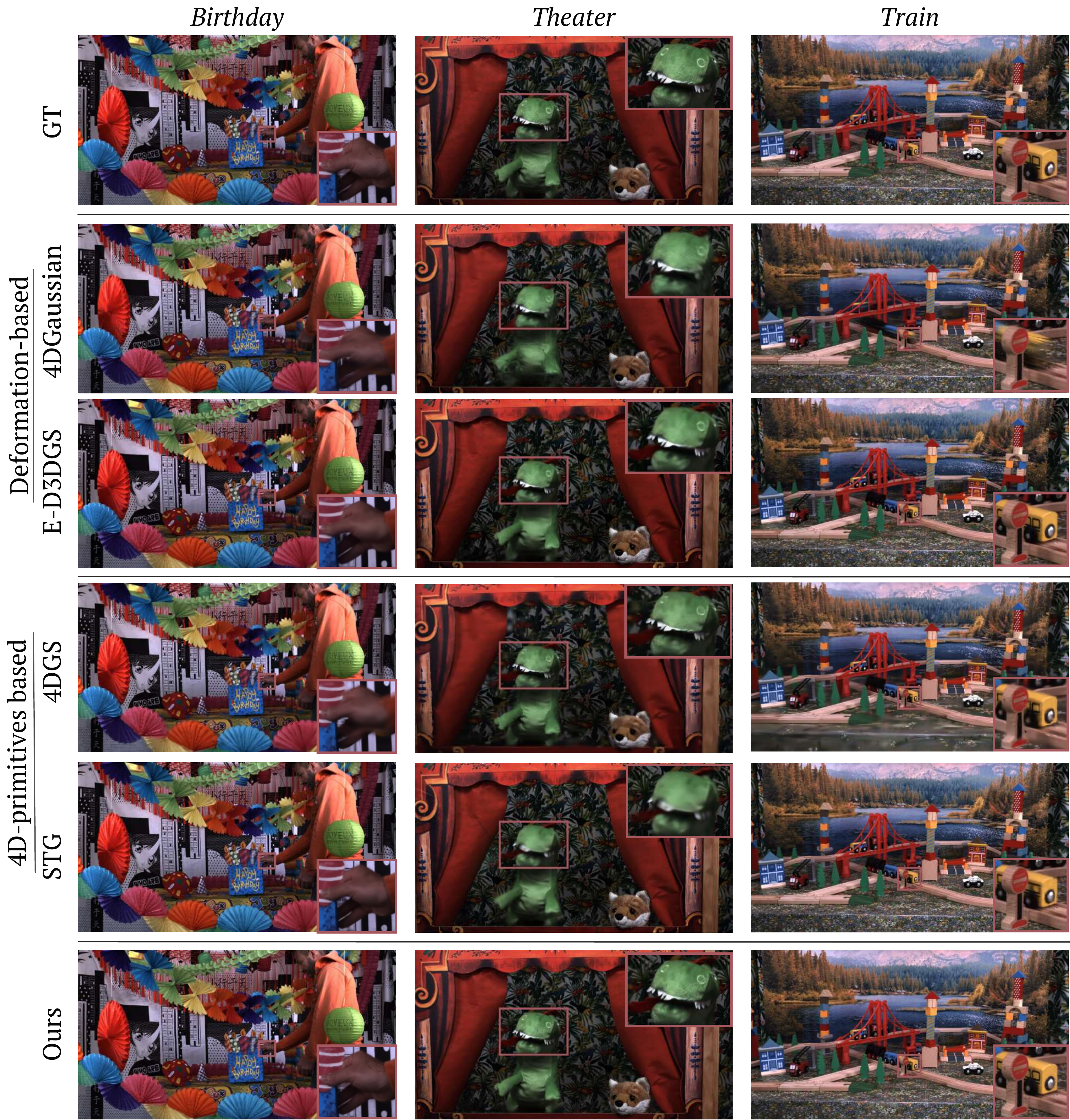}
  \caption{Qualitative comparisons with baseline methods~\cite{wu20244d,yang2023real,li2024spacetime,bae2024ed3dgs} on Technicolor dataset~\cite{sabater2017dataset}. Our method shows consistently superior dynamic details modeling.}
  \label{fig:quali_techno}
\end{figure*}

\begin{table*}[t]
    \begin{center}
    \caption{Quantitative comparisons with SOTA methods on the Technicolor Dataset~\cite{sabater2017dataset}. The \first{best}, \second{second best}, and \third{third best} results are highlighted.}
    \label{table:technicolor}
    \resizebox{0.9\linewidth}{!}{
    \begin{tabular}{l
        cc % Birthday
        cc % Fabien
        cc % Painter
        cc % Theater
        cc % Trains
        | ccc % Mean
        }
    \toprule
     & \multicolumn{2}{c}{Birthday} & \multicolumn{2}{c}{Fabien} & \multicolumn{2}{c}{Painter} & \multicolumn{2}{c}{Theater} & \multicolumn{2}{c}{Train} & \multicolumn{3}{|c}{Mean} \\
    \cmidrule(lr){2-3}\cmidrule(lr){4-5}\cmidrule(lr){6-7}\cmidrule(lr){8-9}\cmidrule(lr){10-11}\cmidrule(lr){12-14}
    Method & \psnr & \dssim & \psnr & \dssim & \psnr & \dssim & \psnr & \dssim & \psnr & \dssim & \psnr & \dssim & \FPS \\
    \midrule
    DyNeRF~\cite{li2022neural} 
        & 29.20 & - & 32.76 & - & 35.95 & - & 29.53 & - & \third{31.58} & - & 31.80 & - & 0.02 \\
    HyperReel~\cite{attal2023hyperreel} 
        & 29.99 & 0.039 & 34.70 & \third{0.053} & \third{35.91} & 0.039 & \second{33.32} & \first{0.053} & 29.74 & \third{0.053} & 32.70 & \third{0.047} & 4.00 \\
            \midrule
   4DGaussians~\cite{wu20244d} 
        & 29.80 & 0.050 & 33.36 & 0.068 & 34.52 & 0.051 & 30.26 & 0.076 & 26.39 & 0.113 & 30.86 & 0.071 & 35 \\
    Def-3DGS~\cite{yang2024deformable} 
        & 30.68 & 0.044 & 33.33 & 0.067 & 34.71 & 0.050 & 29.65 & 0.077 & 26.39 & 0.110 & 30.95 & 0.070 & 76\\
    E-D3DGS~\cite{bae2024ed3dgs} 
        & \first{31.88} & \second{0.033} & \third{34.69} & 0.061 & \second{35.97} & \second{0.036} & 31.04 & 0.064 & 30.87 & \third{0.053} & \third{32.89} & 0.049 & \third{79} \\
            \midrule
    4DGS~\cite{yang2023real} 
        & \third{31.00} & \third{0.038} & 33.57 & 0.058 & {35.73} & 0.042 & \third{31.29} & 0.070 & 28.79 & 0.059 & 32.07 & 0.054 & 55 \\
    STG~\cite{li2024spacetime} 
        & \second{31.65} & \first{0.029} & \second{35.61} & \second{0.047} & {35.73} & \third{0.037} & 31.16 & \second{0.060} & \second{32.61} & \second{0.030} & \second{33.35} & \second{0.040} & \second{86} \\
    \midrule
    \textbf{Ours} 
        & \first{31.88} & \second{0.033} & \first{36.34} & \first{0.046} & \first{36.62} & \first{0.035} & \first{33.49} & \third{0.062} & \first{33.20} & \first{0.026} & \first{34.30} & \first{0.037} & \first{161} \\
    \bottomrule
    \end{tabular}}
    \end{center}

\end{table*}

To evaluate both rendering fidelity and tracking capability, we benchmark Multi4D on three tasks: 
(1) multi-view dynamic novel view synthesis, 
(2) monocular dynamic novel view synthesis, and 
(3) downstream 4D segmentation.

\subsection{Experimental Setup}

\textbf{Datasets.}
We evaluate on three dynamic-scene benchmarks. 
For multi-view reconstruction, we use \textbf{Technicolor}~\cite{sabater2017dataset} (five scenes at $2048 \times 1088$, excluding camera row 2 column 2 following~\cite{attal2023hyperreel}) and \textbf{Neu3D}~\cite{li2022neural} (300-frame sequences at $1352 \times 1014$). 
For monocular reconstruction, we use \textbf{NeRF-DS}~\cite{yan2023nerf}. 
For 4D segmentation, we adopt the \textbf{Neu3D-Mask}~\cite{li2024trase} benchmark. 
All experiments run on a single NVIDIA RTX 4090 GPU.

\textbf{Baselines.}
For novel view synthesis, we compare with NeRF-based methods~\cite{attal2023hyperreel,li2022neural,park2021hypernerf,song2023nerfplayer,yan2023nerf} and Gaussian-based approaches from two domains: deformation-based methods (Def-3DGS~\cite{yang2024deformable}, 4DGaussian~\cite{wu20244d}, DeGauss~\cite{wang2025degauss}, E-D3DGS~\cite{bae2024ed3dgs}) and 4D-primitive-based methods (4DGS~\cite{yang2023real}, STG~\cite{li2024spacetime}). For 4D segmentation, we compare with OpenGaussian~\cite{wu2024opengaussian}, SA4D~\cite{ji2024segment}, and TRASE~\cite{li2024trase}.

\textbf{Metrics.}
For novel view synthesis, we report PSNR$\uparrow$, DSSIM$\downarrow$, and rendering speed (FPS$\uparrow$). 
DSSIM follows the $\text{DSSIM}_1$ formulation with data range 1.0 as in prior work~\cite{li2024spacetime,wu20244d}. 
For 4D segmentation, we report mIoU$\uparrow$ and mAcc$\uparrow$.
\begin{figure*}[t]
  \centering
  \includegraphics[width=0.85\linewidth]{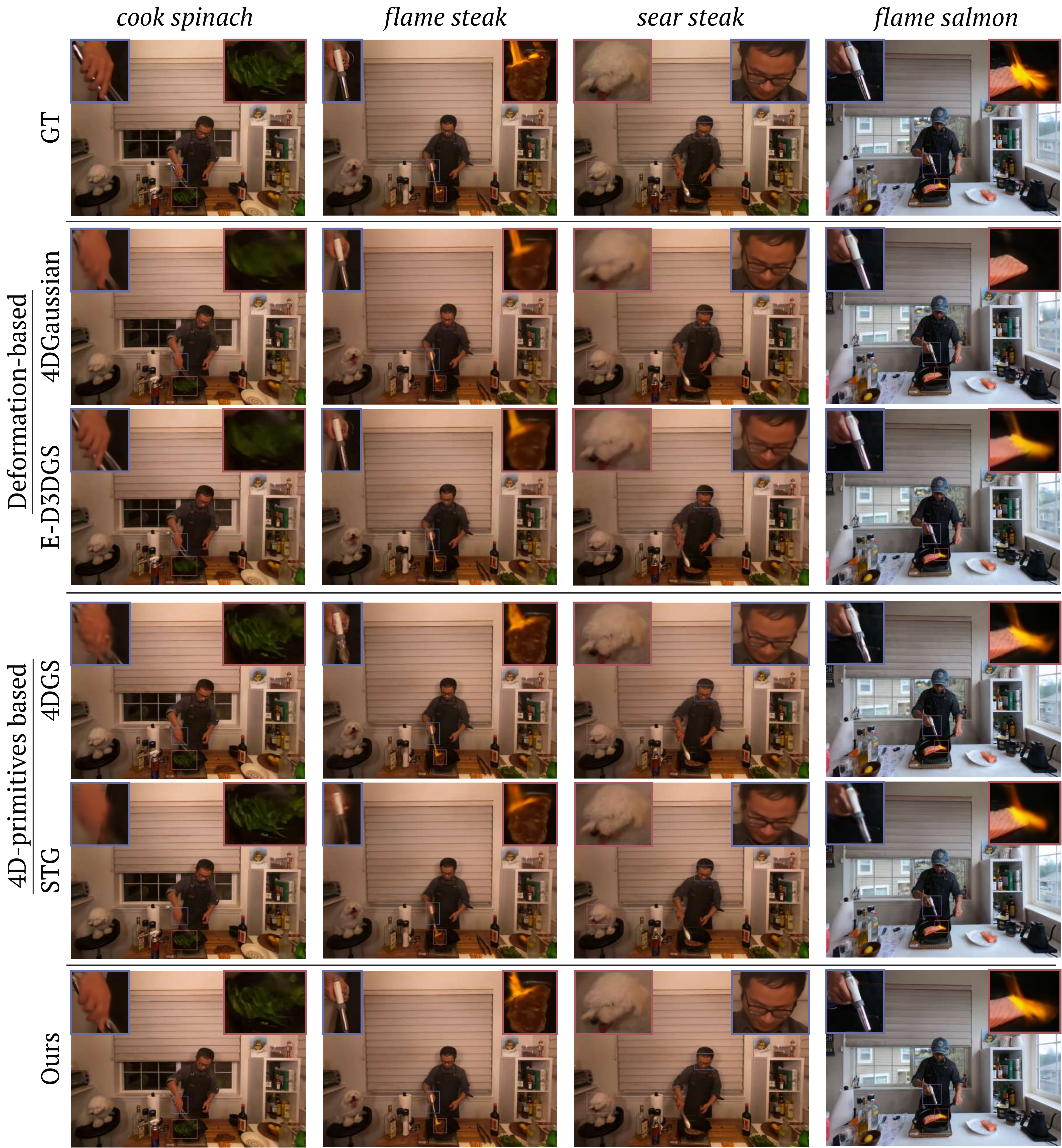}
  \caption{Qualitative comparisons on the Neu3D dataset~\cite{li2022neural}. State-of-the-art deformation-based methods~\cite{wu20244d,bae2024ed3dgs} suffer from over-smoothing in regions with complex appearance changes (highlighted in red), while 4D-primitive approaches~\cite{yang2023real,li2024spacetime} exhibit broken geometry in fast-moving areas (highlighted in blue). Multi4D successfully mitigates both artifacts and consistently achieves the highest visual fidelity. }
  % Multi4D mitigates over-smoothing and geometry fragmentation.
  \label{fig:quali_neu3d}

\end{figure*}

\begin{table*}[t]
    \begin{center}
    \caption{Quantitative comparisons with SOTA methods on the Neu3D dataset~\cite{li2022neural}. The \first{best}, \second{second best}, and \third{third best} results are highlighted. }
    \label{table:n3v}
    \resizebox{0.9\linewidth}{!}{
    \begin{tabular}{l
        cc % Cut Beef
        cc % Cook Spinach
        cc % Sear Steak
        cc % Flame Steak
        cc % Flame Salmon
        cc % Coffee Martini
        | ccc % Mean
        }
    \toprule
     & \multicolumn{2}{c}{Cut Beef} & \multicolumn{2}{c}{Cook Spinach} & \multicolumn{2}{c}{Sear Steak} & \multicolumn{2}{c}{Flame Steak} & \multicolumn{2}{c}{Flame Salmon} & \multicolumn{2}{c}{Coffee Martini} & \multicolumn{3}{|c}{Mean} \\
    \cmidrule(lr){2-3}\cmidrule(lr){4-5}\cmidrule(lr){6-7}\cmidrule(lr){8-9}\cmidrule(lr){10-11}\cmidrule(lr){12-13}\cmidrule(lr){14-16}
    Method & \psnr & \dssim & \psnr & \dssim & \psnr & \dssim & \psnr & \dssim & \psnr & \dssim & \psnr & \dssim & \psnr & \dssim & \FPS \\
    \midrule
    NeRFPlayer$^2$~\cite{song2023nerfplayer}
        & 29.35 & 0.046 & 30.56 & 0.036 & 29.13 & 0.046 & 31.93 & 0.025 & \first{31.65} & \first{0.030} & \first{31.53} & \first{0.025} & 30.69 & 0.034 & 0.05 \\
    HyperReel~\cite{attal2023hyperreel}
        & 32.92 & 0.028 & 32.30 & 0.030 & 32.57 & 0.024 & 32.20 & 0.026 & 28.26 & 0.059 & 28.37 & 0.054 & 31.10 & 0.036 & 2.00 \\
    HexPlane$^{1,2}$~\cite{cao2023hexplane}
        & 32.55 & - & 32.04 & - & 32.39 & - & 32.08 & - & 29.47 & - & - & - & \third{31.71} & - & 0.56 \\
    \midrule
    Def-3DGS~\cite{yang2024deformable}
        & 31.43 & 0.033 & \third{33.06} & 0.027 & 33.01 & 0.024 & 31.83 & 0.025 & 28.70 & 0.043 & 27.88 & 0.047 & 30.98 & 0.033 & 29 \\
    4DGaussian~\cite{wu20244d} 
        & 32.66 & 0.027 & 32.46 & \third{0.026} & 32.49 & 0.022 & 32.75 & \third{0.023} & 29.00 & 0.044 & 27.34 & 0.048 & 31.12 & \third{0.032} & 53 \\
    DeGauss~\cite{wang2025degauss}
        & 32.56 & \second{0.022} & 32.61 & \second{0.025} & 33.20 & 0.022 & 32.75 & \third{0.023} & 29.23 & 0.042 & \third{28.80} & \third{0.042} & 31.52 & \second{0.029} & \second{157} \\
    E-D3DGS~\cite{bae2024ed3dgs}
        & 33.02 & \first{0.021} & 32.71 & \first{0.022} & 31.91 & \third{0.020} & 30.23 & 0.024 & \second{29.79} & \first{0.036} & \second{29.56} & \second{0.032} & 31.20 & \first{0.026} & 70 \\
    \midrule
    4DGS~\cite{yang2023real}
        & \third{33.23} & \third{0.023} & 32.73 & \second{0.025} & \third{33.44} & \third{0.020} & \third{33.19} & \second{0.020} & 28.86 & 0.043 & 27.98 & 0.044 & 31.57 & \second{0.029} & 114 \\
    STG$^2$~\cite{li2024spacetime}
        & \second{33.55} & \first{0.021} & \second{33.18} & \first{0.022} & \second{33.89} & \first{0.017} & \second{33.59} & \first{0.018} & \third{29.48} & \third{0.038} & 28.55 & \third{0.042} & \second{32.04} & \first{0.026} & \third{140} \\
    \midrule
    \textbf{Ours}  
        & \first{34.02} & \first{0.021} & \first{33.30} & \first{0.022} & \first{34.19} & \second{0.018} & \first{34.17} & \first{0.018} & 29.33 & \third{0.038} & \third{28.80} & \third{0.042} & \first{32.30} & \first{0.026} & \first{217}\\
    \bottomrule
    \addlinespace[1ex]
    \multicolumn{16}{l}{\scriptsize $^1$ Excludes the \textit{Coffee Martini} scene} \\
    \multicolumn{16}{l}{\scriptsize $^2$ Due to severe memory overflow, evaluated by training six independent models on 50-frame sequence chunks.} \\
    \end{tabular}}
    \end{center}
\end{table*}
\begin{figure*}[htbp]
  \centering
  \includegraphics[width=0.8\linewidth]{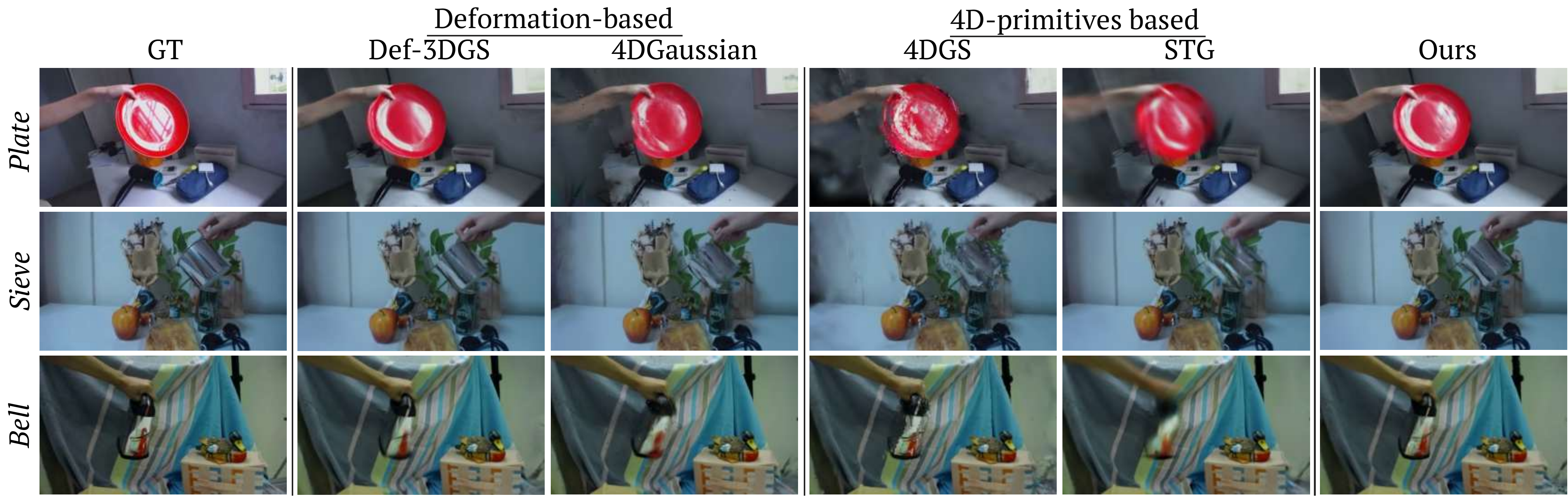}
\caption{Qualitative results on NeRF-DS~\cite{yan2023nerf}. 4D-primitive-based methods~\cite{li2024spacetime,yang2023real} degrade heavily under monocular supervision, while Multi4D preserves coherent geometry and fine specular details via decoupled motion and appearance modeling.}
  \label{fig:quali_nerfds}
\end{figure*}
\begin{table*}[t]
    \setlength{\tabcolsep}{3pt}
    \centering
    \small
    \caption{Quantitative comparison with competitive Nerf methods~\cite{yan2023nerf,park2021hypernerf}, deformation-based gaussian methods~\cite{yang2024deformable,wu20244d}, and 4d primitive-based methods on Monocular NeRF-DS~\cite{yan2023nerf} dataset. The \first{best}, \second{second best}, and \third{third best} results are highlighted.}
    \label{tab:nerfds}
    \resizebox{0.8\textwidth}{!}{%
    \begin{tabular}{l
        cc % As
        cc % Basin
        cc % Bell
        cc % Cup
        cc % Plate
        cc % Press
        cc % Sieve
        | cc % Mean
        }
    \toprule
     & \multicolumn{2}{c}{As} & \multicolumn{2}{c}{Basin} & \multicolumn{2}{c}{Bell} & \multicolumn{2}{c}{Cup} & \multicolumn{2}{c}{Plate} & \multicolumn{2}{c}{Press} & \multicolumn{2}{c}{Sieve} & \multicolumn{2}{|c}{\textbf{Mean}} \\ 
    \cmidrule(lr){2-3} \cmidrule(lr){4-5} \cmidrule(lr){6-7} \cmidrule(lr){8-9} \cmidrule(lr){10-11} \cmidrule(lr){12-13} \cmidrule(lr){14-15} \cmidrule(lr){16-17}
    Method & \psnr & \dssim & \psnr & \dssim & \psnr & \dssim & \psnr & \dssim & \psnr & \dssim & \psnr & \dssim & \psnr & \dssim & \psnr & \dssim \\ 
    \midrule
    NeRF-DS~\cite{yan2023nerf} 
        &\third{25.34} & \second{0.060} & \second{20.23} & \second{0.097} & 22.57 & 0.109 & \first{24.51} & \third{0.060} & \third{19.70} & \second{0.109} & \second{25.35} & \first{0.065} & 24.99 & \first{0.065} & \third{23.24} & \second{0.081} \\
    HyperNeRF~\cite{park2021hypernerf} 
        & 17.59 & 0.074 & \first{22.58} & \first{0.092} & 19.80 & 0.118 & 15.45 & 0.085 & \first{21.22} & \second{0.109} & 16.54 & 0.090 & 19.92 & 0.074 & 19.01 & 0.092 \\
    \midrule

    Def-3DGS~\cite{yang2024deformable} 
        & \second{26.04} & \second{0.060} & \third{19.53} & \third{0.107} & \second{23.96} & 0.103 & \second{24.49} & \second{0.059} & 19.07 & 0.132 & \first{25.52} & \third{0.070} & \second{25.37} & \third{0.069} & \second{23.43} & \third{0.086} \\ 
    4DGaussian~\cite{wu20244d} 
        & 24.85 & \third{0.068} & 19.26 & 0.117 & \third{22.86} & \third{0.099} & 23.82 & 0.065 & 18.77 & \third{0.115} & 24.82 & 0.082 & 25.16 & 0.072 & 22.79 & 0.088 \\
    \midrule
        4DGS~\cite{yang2023real} 
        & 23.44 & 0.089 & 18.66 & 0.135 & 22.84 & 0.106 & 22.21 & 0.082 & 19.06 & 0.123 & 22.52 & 0.118 & 21.87 & 0.108 & 21.51 & 0.108 \\ 
    STG~\cite{li2024spacetime} 
        & 24.57 & 0.072 & 19.17 & 0.121 & 22.82 & \second{0.097} & 23.33 & 0.068 & 17.90 & 0.120 & 24.89 & 0.079 & \third{25.16} & 0.072 & 22.54 & 0.089 \\
    \midrule
    \textbf{Ours} 
        & \first{26.19} & \first{0.058} & 19.50 & \third{0.107} & \first{24.74} & \first{0.081} & \third{24.01} & \first{0.057} & \second{20.36} & \first{0.101} & \third{25.28} & \second{0.069} & \first{25.76 }& \second{0.066} & \first{23.69} & \first{0.077}
        \\ 
    \bottomrule
    \end{tabular}%
    }

\end{table*}

\begin{figure*}[htbp]
  \centering
  \includegraphics[width=0.9\linewidth]{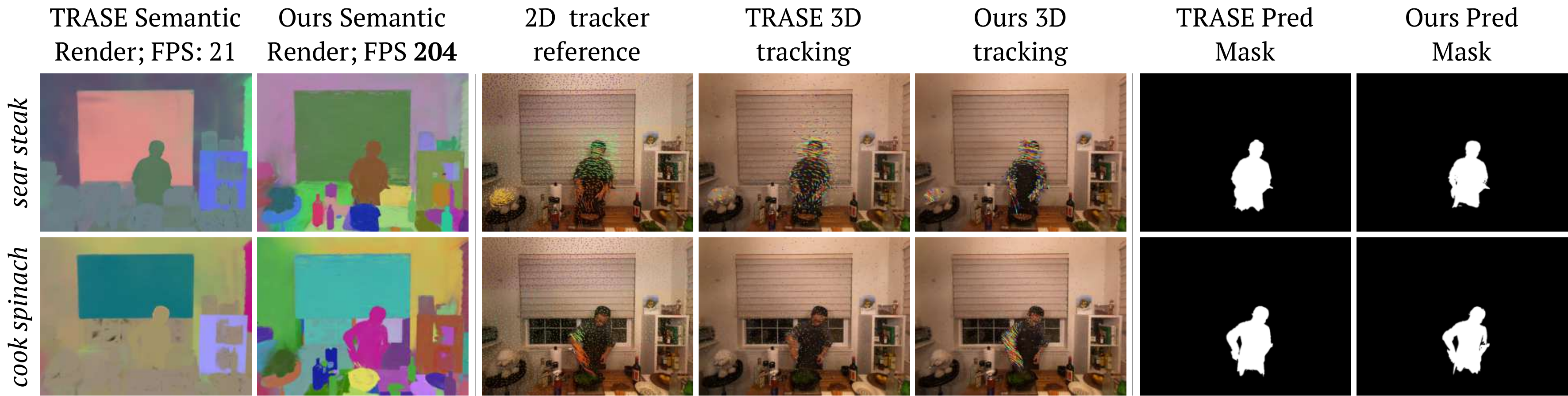}
  \caption{Qualitative Evaluation of 4D Segmentation with TRASE~\cite{li2024trase}. From left to right, we visualize high-dimensional semantic feature renderings, 3D motion trajectories (using Co-tracker~\cite{karaev2024cotracker} as a point-based reference), and predicted 2D object masks.}
  \label{fig:semantic}
\end{figure*}
\begin{table*}[htbp]
  \centering
  \caption{Quantitative results for semantic segmentation on the \textit{Neu3D-Mask} benchmark~\cite{li2024trase}. The \first{best}, \second{second best}, and \third{third best} results are highlighted.}
  \label{tab:neu3d_quantitative}
  \resizebox{0.8\linewidth}{!}{
  \begin{tabular}{l cc cc cc cc cc | cc}
    \toprule
      & \multicolumn{2}{c}{Coffee Martini} & \multicolumn{2}{c}{Cook Spinach} & \multicolumn{2}{c}{Cut Beef} & \multicolumn{2}{c}{Flame Steak} & \multicolumn{2}{c}{Sear Steak} & \multicolumn{2}{|c}{\textbf{Average}} \\  
      \cmidrule(lr){2-3} \cmidrule(lr){4-5} \cmidrule(lr){6-7} \cmidrule(lr){8-9} \cmidrule(lr){10-11} \cmidrule(lr){12-13}
      Method & \mIoU & \mAcc & \mIoU & \mAcc & \mIoU & \mAcc & \mIoU & \mAcc & \mIoU & \mAcc & \mIoU & \mAcc \\
    \midrule
OpenGaussian~\cite{wu2024opengaussian}  
        & 0.8254 & 0.9896 
        & 0.6336 & 0.9798 
        & \second{0.9115} & \second{0.9951} 
        & 0.8199 & 0.9907 
        & \third{0.8986} & \third{0.9943} 
        & 0.8178 & 0.9899 \\

    SA4D~\cite{ji2024segment}               
        & \third{0.8583} & \third{0.9910} 
        & \third{0.8987} & \third{0.9941} 
        & 0.8645 & 0.9914 
        & \first{0.8898} & \first{0.9940} 
        & \second{0.9047} & \second{0.9948} 
        & \third{0.8832} & \third{0.9931} \\
        
    TRASE~\cite{li2024trase}
        & \second{0.9094} & \second{0.9946} 
        & \second{0.9048} & \second{0.9946} 
        & \third{0.8962} & \third{0.9935} 
        & \third{0.8572} & \third{0.9920} 
        & 0.8984 & \third{0.9943} 
        & \second{0.8932} & \second{0.9938} \\
        
    \midrule
    \textbf{Ours}                           
        & \first{0.9108} & \first{0.9947} 
        & \first{0.9161} & \first{0.9954} 
        & \first{0.9424} & \first{0.9967} 
        & \second{0.8698} & \second{0.9929} 
        & \first{0.9321} & \first{0.9963} 
        & \first{0.9142} & \first{0.9952} \\
    \bottomrule
  \end{tabular}
  }
\end{table*}
\subsection{Experimental Results}
We evaluate Multi4D on the Technicolor dataset~\cite{sabater2017dataset}. Quantitative and qualitative results are shown in \cref{table:technicolor,fig:quali_techno}. Our method improves PSNR by 0.95\,dB over the strongest Gaussian baseline~\cite{li2024spacetime,yang2023real,bae2024ed3dgs} while achieving real-time rendering at 161 FPS. Multi4D also achieves state-of-the-art results on Neu3D with superior reconstruction quality and 217 FPS, as shown in \cref{fig:quali_neu3d} and \cref{table:n3v}.
As illustrated in \cref{fig:quali_techno,fig:quali_neu3d}, Multi4D better preserves high-frequency dynamic details, whereas deformation-based methods exhibit temporal blurring and 4D-primitive approaches produce fragmented geometry.

We further evaluate Multi4D on the monocular NeRF-DS dataset~\cite{yan2023nerf}, which exposes limitations of 4D-primitive frameworks~\cite{li2024spacetime,yang2023real} under sparse supervision. As shown in \cref{tab:nerfds}, these methods lacking a holistic motion prior degrade significantly and often produce floating artifacts (see \cref{fig:quali_nerfds}). In contrast, Multi4D leverages persistent Gaussians $\mathcal{G}_d$ for coherent motion modeling and transient Gaussians $\mathcal{G}_t$ for localized specular highlights, resulting in consistently stronger reconstructions.

Multi4D also extends naturally to 4D Segmentation, achieving state-of-the-art $0.9142$ mIoU on the Neu3D-Mask benchmark (\cref{tab:neu3d_quantitative}). By restricting semantic optimization to the persistent subset $\mathcal{G}_p$, we avoid transient appearance noise while preserving temporal identity. As shown in \cref{fig:semantic}, Multi4D produces sharper masks and more stable tracks than TRASE~\cite{li2024trase}. Our representation uses only 13k dynamic Gaussians (vs. 624k in TRASE), enabling 32-dimensional feature rendering at 204 FPS—nearly a $10\times$ speedup over the 21 FPS baseline.

Additional per-component renderings and qualitative novel view synthesis results are provided in the supplementary material.
\section{Ablation Study \& Efficiency Analysis}
\begin{table}[htbp]
    \centering
    \caption{Ablation of core components on Neu3D dataset~\cite{li2022neural} (avg. over 4 scenes).}
    \label{tab:ablation}
    \resizebox{0.65\linewidth}{!}{
    \begin{tabular}{l | cc | cc}
    \toprule
    Configuration & \psnr & \dssim & \scalebox{0.8}{Dyn. GS \#$^1$ $\downarrow$} & \scalebox{0.8}{Storage$^2$ $\downarrow$} \\
    \midrule
    Baseline 4DGS~\cite{yang2023real}
        & 33.14 & 0.0219 & \textcolor{red}{4215 k} & 2.6 GB \\
                \midrule
    w/o $\mathcal{G}_d$ (Persistent dynamic subset)
        & 32.78 & 0.0237 & \textcolor{red}{1139 k} & 727.5 MB \\
    w/o $\mathcal{G}_t$ (Transient dynamic subset)
        & 32.86 & 0.0217 & \textcolor{blue}{25 k} & 105.4 MB \\
        \midrule
    w/o Periodical Lifting (Sec.~\ref{sec:adaptive_lift}) 
        & 33.22 & 0.0216 & (\textcolor{blue}{13 k} + \textcolor{red}{132 k}) & 184.84 MB \\
    w/o $\mathcal{L}_\text{diversity}$ (Sec.~\ref{sec:hybrid_render}) 
        & 33.66 & 0.0203 & (\textcolor{blue}{19 k} + \textcolor{red}{257 k}) & 263.8 MB \\
    w/o Mask-Aware Pruning (Sec.~\ref{sec:pruning}) 
        & 33.68 & 0.0199 & (\textcolor{blue}{70 k} + \textcolor{red}{659 k}) & 527.9 MB \\
    \midrule
    \textbf{Multi4D (Full)} 
        & \textbf{33.92} & \textbf{0.0197} & ({\textcolor{blue}{13 k} + \textcolor{red}{152 k}}) & {214.7 MB} \\
    \bottomrule
    \addlinespace[1ex]
    \multicolumn{5}{l}{\scriptsize 1 Dyn. GS \#: Dynamic Gaussians number, denoted as (\textcolor{blue}{Persistent $\mathcal{G}_d$} + \textcolor{red}{Transient $\mathcal{G}_t$}).} \\
    \multicolumn{5}{l}{\scriptsize 2 Storage denotes the combined size of the three Gaussian subsets and the deformation network.} \\
    \end{tabular}
    }

\end{table}
\textit{Ablation Study.}
We evaluate the core components of \textit{Multi4D} on four Neu3D scenes (\textit{Cut Beef}, \textit{Cook Spinach}, \textit{Sear Steak}, \textit{Flame Steak}), summarized in \cref{tab:ablation}.

\textbf{Role of Dynamic Subsets $\mathcal{G}_d$, $\mathcal{G}_t$.}
Removing the persistent dynamic subset ($\mathcal{G}_d$) degrades reconstruction quality (32.78 PSNR), as motion must be approximated by randomly initialized transient primitives. Without stable structural initialization, accurate 4D modeling becomes difficult, leading to fragmented geometry and excessive primitive growth. Conversely, removing the transient subset ($\mathcal{G}_t$) limits high-frequency appearance modeling, reducing PSNR to 32.86 despite a compact representation (25k dynamic Gaussians). This confirms that $\mathcal{G}_d$ models coherent motion while $\mathcal{G}_t$ captures residual appearance changes.

\textbf{Velocity-Aware Periodical Lifting}. Replacing velocity-aware lifting with random initialization reduces PSNR by 0.70\,dB (33.92 → 33.22), as transient primitives lack inherited motion priors and under-model high-frequency dynamics.

\textbf{Diversity Loss.}
Removing $\mathcal{L}_{\text{diversity}}$ weakens subset specialization, causing redundant modeling across persistent and transient sets. This increases dynamic primitives by 67\% (276k vs.\ 165k) and enlarges storage while reducing fidelity.

\textbf{Mask-Aware Pruning.}
Without the mask-aware utility-based pruning, temporal over-parameterization re-emerges: the modeled dynamic Gaussian number grows to 729k and storage increases by 145\%, while unregularized Gaussians degrade rendering quality.

\textit{Efficiency and Compactness.}
Compared to 4DGS~\cite{yang2023real}, \textit{Multi4D} uses only 165k dynamic Gaussians—a $25\times$ reduction from 4.2M—while improving PSNR to 33.92. Model size drops from 2.6\,GB to 214.7\,MB, and training converges in 1.2 hours versus 5.5 hours on a single RTX 4090 ($4.6\times$ faster).
\section{Discussion and Conclusion}

We presented \textit{Multi4D}, a multi-level framework for dynamic 3D Gaussian Splatting that resolves the tension between motion consistency and photometric fidelity. By competitively allocating modeling capacity across static, persistent, and transient Gaussian subsets under a unified objective, Multi4D enables residual-driven specialization instead of monolithic representations, suppressing motion over-factorization and temporal over-parameterization to produce compact, high-fidelity reconstructions. This structured decomposition preserves coherent motion while capturing high-frequency appearance variation, and performing semantic reasoning on the persistent subset further enables efficient and highly accurate downstream 4D Segmentation.

\textit{Limitations and Future Work.}
Although Multi4D substantially reduces the number of dynamic primitives through optimization-driven compactness, it does not currently incorporate explicit attribute compression. Future work could explore post-training deformation distillation, Gaussian quantization, or lightweight deformation parameterizations to further translate structural compactness into improved storage efficiency. 
%\
% \clearpage
% \newpage

\bibliographystyle{splncs04}
\bibliography{main}

\clearpage
\appendix
\section{Supplementary Material}
\subsection{Overview of the Multi4D Pipeline}

For convenience, we include the multi-level competitive allocation pipeline figure of Multi4D from the main paper to facilitate referencing the subcomponents described throughout the supplementary material.

\begin{figure*}[h]
\centering
\includegraphics[width=\linewidth]{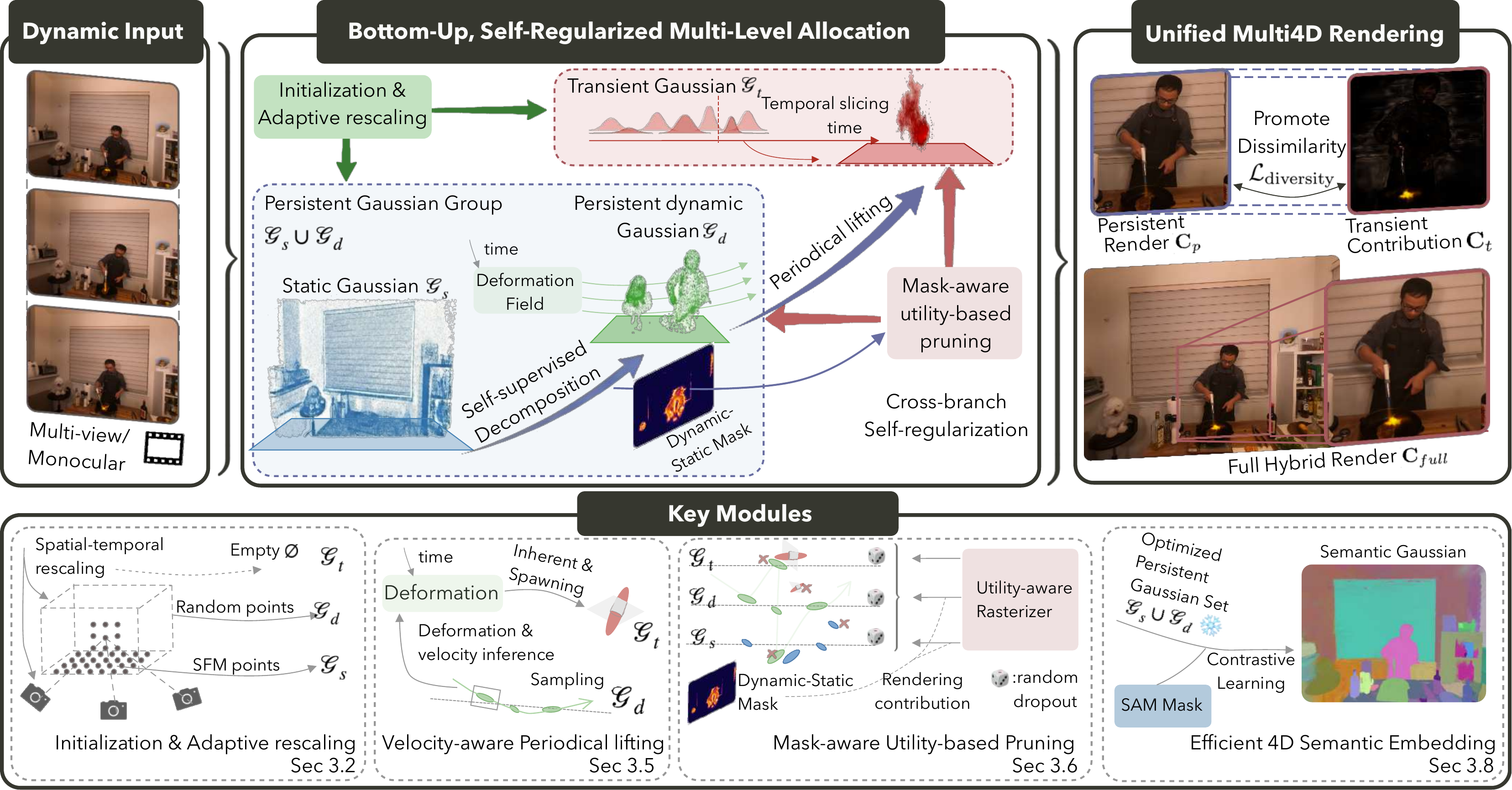}
\caption{Overview of the Multi4D framework. 
The scene is decomposed into three Gaussian subsets: static ($\mathcal{G}_s$), persistent dynamic ($\mathcal{G}_d$), and transient ($\mathcal{G}_t$). 
These subsets are jointly optimized under a bottom-up, self-regularized multi-level allocation scheme through a shared renderer, while maintaining independent optimization and density-control dynamics. 
This design enables competitive allocation and emergent subset specialization. 
Numbers in the figure correspond to the section numbers in the main paper describing each core component.}
\label{fig:pipeline_supp_app}
\end{figure*}
\section{Analysis on Subset Specialization}

\begin{figure*}[htbp]
  \centering
  \includegraphics[width=\linewidth]{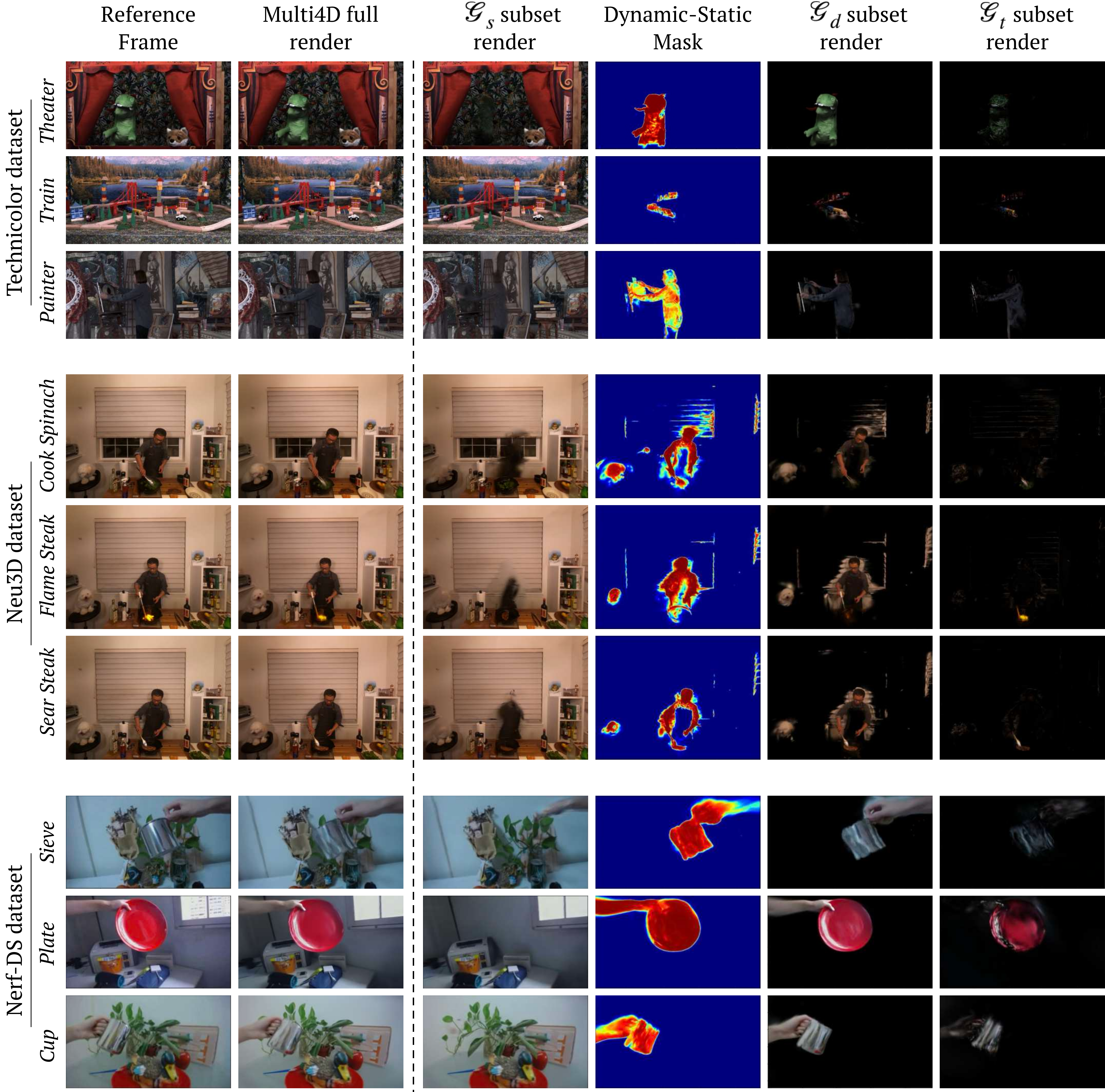}
\caption{Subset specialization produced by our multi-level competitive allocation strategy. 
For each scene we visualize the ground truth, the full Multi4D render, individual renders of each subset, and the learned dynamic–static separation mask. 
The static subset ($\mathcal{G}_s$) reconstructs the background geometry, persistent Gaussians ($\mathcal{G}_d$) maintain coherent dynamic structure and motion identity, while transient primitives ($\mathcal{G}_t$) capture localized high-frequency residuals such as fine appearance changes or specular highlights.
Across Neu3D~\cite{li2022neural}, Technicolor~\cite{sabater2017dataset}, and NeRF-DS~\cite{yan2023nerf}, this multi-level subset specialization jointly models dynamic scenes, producing compact representations while preserving fine dynamic details, achieving consistently superior performance.}
% \vspace{-7em}
  \label{fig:subset}
\end{figure*}
To empirically validate the effectiveness of our multi-level competitive allocation paradigm, we visualize the learned subset specialization across three diverse dynamic view synthesis benchmarks: Neu3D~\cite{li2022neural} (complex appearance variations and articulated motion), Technicolor~\cite{sabater2017dataset} (high-fidelity multi-view captures with rich texture details), and NeRF-DS~\cite{yan2023nerf} (challenging monocular sequences with strong specularities and limited geometric constraints).

As shown in \cref{fig:subset}, we present the ground truth, the full Multi4D render, individual renders of each Gaussian subset, and the learned dynamic–static separation mask. This visualization provides a direct breakdown of how each subset contributes to the final reconstruction. The static subset ($\mathcal{G}_s$) consistently captures time-invariant scene structure, forming a stable geometric backbone. The persistent dynamic subset ($\mathcal{G}_d$) models coherent object motion and preserves long-term geometric identity through the deformation field. Meanwhile, the transient subset ($\mathcal{G}_t$) activates only where photometric residuals cannot be explained by geometric deformation alone, capturing high-frequency appearance variations, such as fire, smoke and specular highlights.

This emergent specialization arises naturally from the competitive allocation mechanism. Because all subsets are jointly optimized through a shared renderer, each subset competes to explain photometric residuals while respecting its inductive bias. At the same time, each subset follows its own optimization and densification dynamics; once a subset adequately explains a region, redundant modeling in the others is naturally suppressed. Following a bottom-up strategy, where subsets are initialized inversely proportional to their representational capacity, the three subsets are progressively optimized under cross-level self-regularization, resulting in high-quality yet highly compact scene representations.

As a result, the representation avoids two common failure modes of existing dynamic Gaussian approaches: motion over-factorization in deformation-based models and temporal over-parameterization in 4D primitive methods. Instead, Multi4D distributes modeling capacity across complementary regimes, yielding compact representations that preserve coherent geometry while capturing high-frequency dynamics. This structured specialization directly contributes to the improved rendering fidelity, reduced primitive counts, and higher runtime efficiency reported in the main experiments.

\section{Detailed Implementation and Methodology}
\subsection{Independent Optimization and Densification per Subset}
In standard dynamic 3DGS pipelines, densification and optimization are applied globally across all primitives. In Multi4D, we instead manage each Gaussian subset independently. This allows the different subsets to specialize and model fine details with cross-subset self-regularization.

Concretely, each subset ($\mathcal{G}_s$, $\mathcal{G}_d$, $\mathcal{G}_t$) has its own Adam optimizer, learning rate schedule, and densification–pruning policy. As a result, each subset can grow and adapt according to its own role in the representation.

\noindent \textbf{1. Static Background ($\mathcal{G}_s$).} 
The static subset models the time-invariant environment. Its optimizer updates only the canonical Gaussian parameters. Densification is triggered during Phase I using 2D screen-space positional gradients ($\nabla_{2D}\mu_s$). Because the background spans the largest spatial extent, this subset is allocated a larger primitive budget with relatively relaxed pruning.

\noindent \textbf{2. Persistent Dynamics ($\mathcal{G}_d$).} 
This subset models the geometry of moving actors. Its optimizer updates both the Gaussian parameters and the HexPlane-based~\cite{cao2023hexplane,wu20244d} deformation network $\Phi_g$. Densification is driven by gradients accumulated by the \emph{deformed} Gaussians. However, the actual clone/split operations are applied to the canonical Gaussians in practice~\cite{wu20244d,yang2024deformable}. This simplification provides additional regularization and improves training efficiency, though it makes modeling fine details more challenging. In our framework, we exploit this property to encourage specialization between the persistent subset ($\mathcal{G}_d$) and the transient subset ($\mathcal{G}_t$).

\noindent \textbf{3. Transient Primitives ($\mathcal{G}_t$).} 
The transient subset captures high-frequency appearance changes. Because these 4D primitives are highly expressive, their capacity is tightly controlled. Densification uses both spatial and temporal gradients ($\nabla_t\mu_t$), while progressive opacity pruning limits excessive primitive growth and maintains a compact representation.

This isolated optimization strategy encourages the transient subset to remain sparse, allowing new primitives to emerge only in regions where the physically constrained persistent models ($\mathcal{G}_s \cup \mathcal{G}_d$) cannot explain the observed appearance changes.

\subsection{Training Schedule and Optimization Stages}
\label{supple:stages}
Multi4D adopts a progressive optimization schedule that gradually encourages subset specialization while avoiding degenerate geometry.

\noindent \textbf{Stage I: Subset Formation (0 -- $T_{\text{sep}}$).} \\
To prevent the model from introducing non-physical motion during early optimization, the deformation module ($\Phi_g$) is frozen for the first 2k iterations. During this period, training focuses on establishing stable canonical geometry for the static ($\mathcal{G}_s$) and persistent dynamic ($\mathcal{G}_d$) subsets.

After 2k iterations, the deformation module is activated, allowing $\mathcal{G}_d$ to model coherent motion. During this stage, the full set of decoupling losses (including $\mathcal{L}_{\text{diversity}}$ and dynamic–static compositing) is applied. Once dynamic--static separation is sufficiently stable, velocity-aware lifting periodically spawns Gaussians into the transient subset ($\mathcal{G}_t$). To maintain compactness, utility-based pruning is applied separately to each subset.

\noindent \textbf{Stage II: Unified Rendering Refinement ($T_{\text{sep}}$ -- $T_\text{end}$).} \\
After subset specialization stabilizes, the explicit dynamic–static separation process and $\mathcal{L}_{\text{sep}}$ are disabled, and training proceeds using the unified depth-sorted hybrid renderer. Optimization then focuses on improving photometric fidelity while preserving the established compact geometric structure. The late-stage objective keeps the photometric term, depth-ordering regularization, persistent render supervision, and depth smoothness on the persistent path.

\noindent \textbf{Densification and Pruning Dynamics.} \\
Across all active subsets, Gaussians are cloned or split every 100 iterations when positional gradients exceed predefined thresholds, following the standard 3DGS densification procedure. Opacities are periodically and partially reset to stabilize the foreground–background decomposition~\cite{wang2025degauss} and promote compact representations. Because transient primitives ($\mathcal{G}_t$) are highly expressive 4D elements, progressive utility-based pruning based on their final rendering contribution is applied to limit excessive primitive growth.
\subsection{Self-Supervised Dynamic–Static Decomposition Details}
\label{supple:separation_detail}
\begin{figure*}[t]
  \centering
  \includegraphics[width=0.9\linewidth]{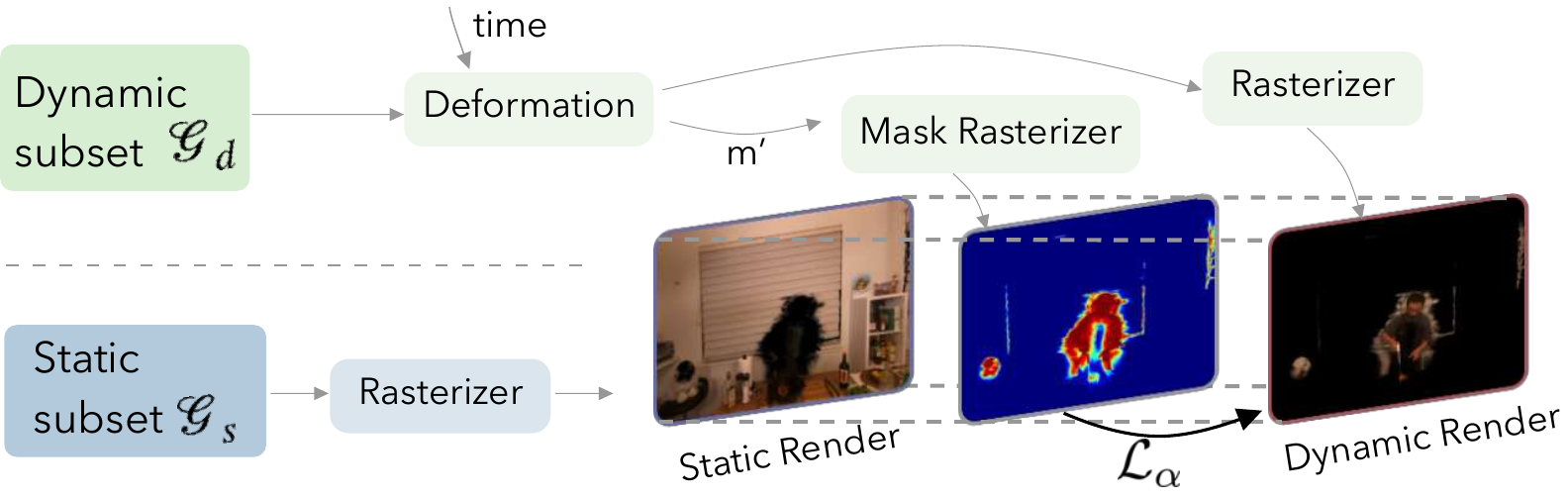}
\caption{Self-supervised dynamic–static decomposition. 
A learned dynamic probability mask separates moving actors from the static background without external supervision. 
An additional opacity regularization ($\mathcal{L}_{\alpha}$) aligns the rendered dynamic alpha with the mask, suppressing dynamic opacity in static regions and enabling structural decoupling between persistent dynamics and static geometry during early training.}
\label{figsupple:dynasta}
\end{figure*}

During the Canonical Initialization and Subset Specialization stages (Phase I), we enforce a structural separation between persistent dynamic actors ($\mathcal{G}_d$) and the static background ($\mathcal{G}_s$). Following the probabilistic masking formulation of DeGauss~\cite{wang2025degauss}, this decomposition is learned in a fully self-supervised manner through asymmetric initialization and mask-guided supervision, with overall pipeline shown in ~\cref{figsupple:dynasta}.

\noindent \textbf{1. Mask Rendering Mechanism.} \\
Each persistent dynamic Gaussian $g_i \in \mathcal{G}_d$ is augmented with a base mask logit $m_i$. To account for deformation-dependent topology changes, a lightweight MLP $\mathcal{D}_m$ predicts a time-dependent offset:
\begin{equation}
m'_i(t) = m_i + \mathcal{D}_m\big(\mathcal{H}(\mu_i, t)\big)
\end{equation}
where $\mathcal{H}(\mu_i,t)$ denotes the spatiotemporal features extracted from the HexPlane deformation module. During Phase I, the spherical harmonic color of $\mathcal{G}_d$ is replaced with the activated mask probability $\sigma(m'_i(t))$. Rasterization produces a dynamic probability map $M_d \in [0,1]^{H\times W}$, and the static mask is defined as $M_s = 1 - M_d$.

\noindent \textbf{2. Composite Rendering.} \\
To supervise the decomposition, we render the dynamic foreground ($C_d$) and static background ($C_s$) separately and combine them using the predicted mask:
\begin{equation}
C_{comp} = M_d \odot C_d + (1 - M_d) \odot C_s .
\end{equation}

\noindent \textbf{3. Separation Loss ($\mathcal{L}_{sep}$).} \\
The dynamic–static decomposition is enforced through a combined loss $\mathcal{L}_{sep}$ that aggregates the compositing and regional supervision terms:
\begin{equation}
\mathcal{L}_{sep} =
\lambda_{comp}\|C_{comp}- \gamma C_{gt}\|_1
+
\mathcal{L}_{regional},
\end{equation}
where $\gamma = 0.9$ during the first 2000 iterations and $\gamma = 1$ afterwards. 
This slightly attenuates the target signal in early training to prevent the high-capacity foreground model from overfitting background regions before the static geometry stabilizes.
Regional supervision further discourages subsets from modeling regions outside their assignments:
\begin{equation}
\begin{aligned}
\mathcal{L}_{regional} &=
\| (C_d \odot \mathbb{I}_{M_d>\tau_{vis}}) -
   (C_{gt} \odot \mathbb{I}_{M_d>\tau_{vis}}) \|_1 \\
&\quad +
\| (C_s \odot \mathbb{I}_{M_s>\tau_{vis}}) -
   (C_{gt} \odot \mathbb{I}_{M_s>\tau_{vis}}) \|_1 ,
\end{aligned}
\end{equation}
where $\tau_{vis}=0.49$ and $\mathbb{I}$ denotes the indicator function.

Because $\mathcal{G}_s$ is densely initialized from COLMAP points while $\mathcal{G}_d$ begins from sparse random initialization (10k points), the optimization naturally assigns stable regions to the static subset. As training progresses, the dynamic mask $M_d$ contracts around moving actors, isolating physical motion without external supervision.
\subsection{Exact Loss Formulations}

Our optimization objective combines photometric supervision with structural regularization to enforce subset specialization and geometric consistency. The overall loss is defined as:
\begin{equation}
\mathcal{L}_{total} =
\mathcal{L}_{color}
+
\mathcal{L}_{sep}
+
\mathcal{L}_{diversity}
+
\mathcal{L}_{reg}.
\end{equation}

\noindent \textbf{1. Photometric Supervision ($\mathcal{L}_{color}$).} \\
The base rendering loss combines $L_1$ error and SSIM between the rendered image $C$ and the ground-truth image $C_{gt}$:
\begin{equation}
\mathcal{L}_{color} =
\lambda_{L1}\|C - C_{gt}\|_1
+
\lambda_{SSIM}(1-\text{SSIM}(C,C_{gt})).
\end{equation}

During the early subset formation stage, this loss is applied not only to the full hybrid render but also independently to the persistent foreground and static background renders to encourage dynamic–static separation. We set $\lambda_{L1}=1.0$ and $\lambda_{SSIM}=0.4\times\lambda_{downscale\_ulti}$.

\noindent \textbf{2. Dynamic–Static Separation Loss ($\mathcal{L}_{sep}$).} \\
To enable self-supervised decomposition between persistent dynamics and static background, we apply the separation objective described in~\cref{supple:separation_detail}. This loss aggregates the mask-guided compositing and regional supervision terms that enforce structural decoupling between $\mathcal{G}_d$ and $\mathcal{G}_s$ during the subset formation stage.

\noindent \textbf{3. Cross-Subset Diversity Loss ($\mathcal{L}_{diversity}$).} \\
To encourage regime specialization and prevent redundant modeling, we penalize structural similarity between outputs of competing subsets. In particular, we discourage transient primitives from reproducing persistent geometry using a mask-weighted SSIM penalty:
\begin{equation}
\mathcal{L}_{diversity} =
\lambda_{div}
\frac{1}{|\mathcal{P}|}
\sum_{u\in\mathcal{P}}
\text{SSIM}(C_{4D}(u),C_{3D}(u))
\cdot
\alpha_{hybrid}(u),
\end{equation}
where $\alpha_{hybrid}$ denotes the rendered opacity mask restricting the penalty to valid foreground regions. We use $\lambda_{div}=0.1$. A weaker diversity penalty is also applied between the dynamic and static renders to encourage clean foreground–background separation.

\noindent \textbf{4. Geometric Regularization ($\mathcal{L}_{reg}$).} \\
The regularization term aggregates several constraints that maintain geometric stability and enforce physically plausible rendering:

\begin{itemize}
\item \textbf{Mask-Aware Opacity Regularization.}  
To discourage persistent dynamic Gaussians from occupying static regions, we supervise the rendered alpha map of the dynamic subset using the binarized dynamic–static separation mask as a target. Let $\alpha_d$ denote the rendered opacity of $\mathcal{G}_d$, and let $\mathbb{I}_{M_d>\tau}$ be the binary mask obtained by thresholding the dynamic probability map. We define
\begin{equation}
\mathcal{L}_{\alpha} =
\lambda_{\alpha}
\big\|
\alpha_d - \mathbb{I}_{M_d>\tau}
\big\|_1 ,
\end{equation}
This suppresses dynamic opacity in static regions and improves separation.

\item \textbf{Depth Ordering.}  
Transient primitives modeling effects such as fire or specular highlights should lie on or in front of persistent geometry. We therefore apply a one-sided depth constraint:
\begin{equation}
\mathcal{L}_{depth} =
\lambda_{depth}
\frac{1}{|\mathcal{P}|}
\sum_{u\in\mathcal{P}}
\max(0, D_{4D}(u)-D_{3D}(u)).
\end{equation}

\item \textbf{Scale Regularization.}  
To prevent excessively large primitives, Gaussians whose maximum scale $s_{max}$ exceeds $10\%$ of the scene extent $\mathcal{E}_{cam}$ are penalized:
\[
\mathcal{L}_{scale} =
0.01\cdot\max(0, s_{max}-0.1\mathcal{E}_{cam}).
\]

\item \textbf{Aspect Ratio Regularization.}  
To avoid degenerate elongated primitives:
\[
\mathcal{L}_{aspect} =
0.1\cdot
\max\!\left(0,\frac{s_{max}}{s_{min}}-5\right).
\]

\item \textbf{Depth Smoothness.}  
An edge-aware total variation penalty is applied to the rendered dynamic depth map to suppress floating artifacts:
\begin{equation}
\begin{aligned}
\mathcal{L}_{TV} =
\lambda_{TV}\frac{1}{|\mathcal{P}|}
\sum_{u\in\mathcal{P}}
\Big(
|\nabla_x D_d(u)|e^{-\|\nabla_x C_{gt}(u)\|_1} \\
+
|\nabla_y D_d(u)|e^{-\|\nabla_y C_{gt}(u)\|_1}
\Big).
\end{aligned}
\end{equation}
\end{itemize}

Finally, for the persistent dynamic subset ($\mathcal{G}_d$), temporal smoothness is enforced by applying a total variation penalty directly to the HexPlane deformation grids.

All these terms are jointly aggregated into the regularization term $\mathcal{L}_{reg}$.
\subsection{Implementation Details}
All experiments are implemented in PyTorch and trained on a single RTX 4090 GPU. Training runs for $20{,}000$ iterations. The canonical initialization stage lasts for $T_{\text{init}}=2{,}000$ iterations, during which the deformation module is frozen. The model transitions from subset formation to unified rendering refinement at $T_{\text{sep}}=10{,}000$ iterations.

Velocity-aware lifting starts at iteration $6{,}000$ and is performed every $50$ iterations until $10{,}000$, sampling at most $K=2000$ activated persistent Gaussians with dynamic-score threshold $\tau=0.05$. Densification is active from iterations $500$ to $15{,}000$: the dynamic subset $\mathcal{G}_d$ is densified until $10{,}000$, while the static and transient subsets continue densification until $15{,}000$. Gaussian cloning and splitting are evaluated every $100$ iterations following the standard 3DGS procedure. For monocular datasets, transient densification is disabled and transient primitives are introduced only through velocity-aware lifting.

For $\mathcal{G}_d$, spatiotemporal features are represented using a HexPlane grid with resolution $[64,64,64,150]$. The deformation module is a lightweight MLP with one hidden layer of width $128$. We apply temporal smoothness, time-plane sparsity, and plane-TV regularization with weights $1.0\times10^{-3}$, $1.0\times10^{-4}$, and $2.0\times10^{-4}$, respectively. The release configuration further uses soft velocity regularization with weight $0.005$, depth-ordering loss with weight $0.01$, foreground alpha compactness with weight $0.01$, and edge-aware depth TV with weight $0.01$.

All parameters are optimized with Adam. We use exponentially decayed learning rates for Gaussian positions $(1.6\times10^{-4}\rightarrow1.6\times10^{-6})$, the deformation MLP $(1.6\times10^{-4}\rightarrow1.6\times10^{-5})$, and HexPlane grids $(8.0\times10^{-4}\rightarrow5.0\times10^{-6})$. Fixed learning rates are used for spherical harmonics $(2.5\times10^{-3})$, opacity $(5.0\times10^{-2})$, scaling $(5.0\times10^{-3})$, and rotation $(1.0\times10^{-3})$. The released code includes the exact configuration files used for all reported experiments.

\section{Downstream 4D segmentation}
\begin{figure*}[htbp]
  \centering
  \includegraphics[width=0.9\linewidth]{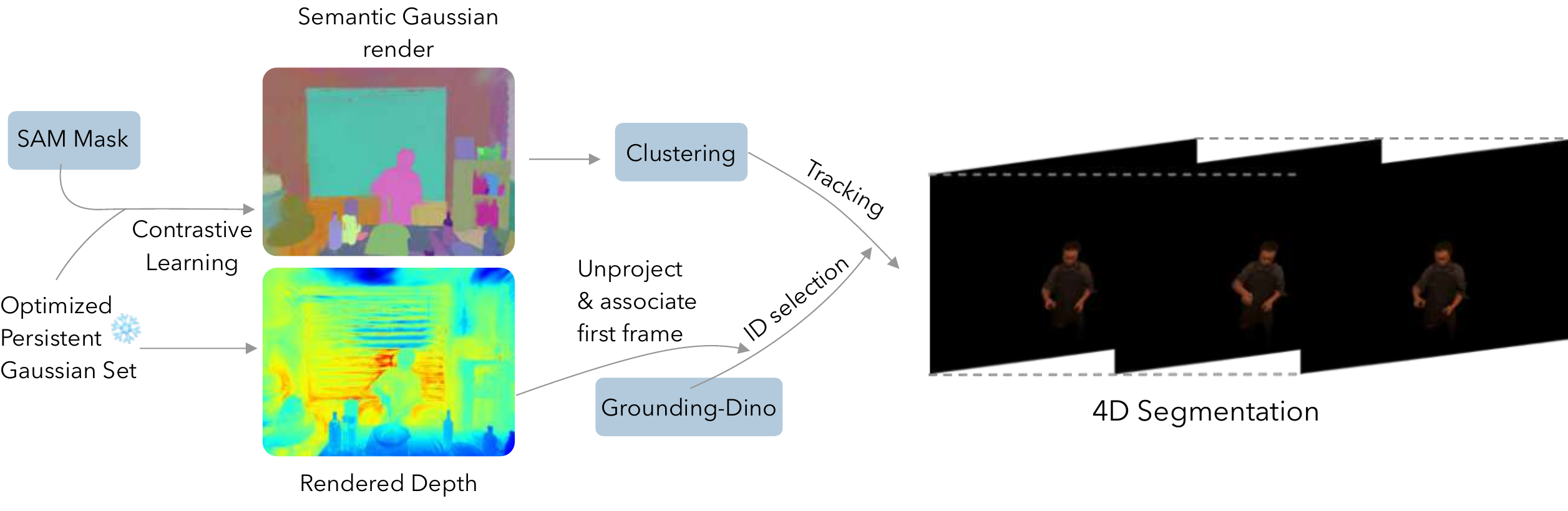}
\caption{4D semantic segmentation pipeline. 
After dynamic reconstruction, semantic features are learned on the optimized persistent Gaussian set using SAM masks and contrastive learning. 
The rendered semantic features are clustered to obtain consistent object identities. 
At inference, a text prompt (Grounding-DINO) or mask is associated with the first frame via depth unprojection, and the selected cluster is tracked across time through the deformation field to produce temporally consistent 4D segmentation.}
  \label{fig:seg4d_supple}
\end{figure*}

\subsection{4D Semantic Tracking}

To evaluate the structural consistency of our decomposed representation, we apply Multi4D to downstream 4D semantic tracking. Following the mask-driven contrastive learning framework of TRASE~\cite{li2024trase}, we optimize semantic features on the reconstructed scene. We perform semantic optimization exclusively on the persistent geometry
\[
\mathcal{G}_p = \mathcal{G}_s \cup \mathcal{G}_d,
\]
while the transient subset $\mathcal{G}_t$ is discarded. This removes short-lived appearance primitives that break temporal identity.

After dynamic reconstruction converges, all geometric parameters and the HexPlane deformation network $\Phi_g$ are frozen. Each persistent Gaussian is augmented with a learnable semantic feature vector $f_i\in\mathbb{R}^{32}$ initialized randomly and optimized using Adam with learning rate $0.0025$. The overall pipeline is shown in~\cref{fig:seg4d_supple}.

During training, the frozen deformation field maps persistent Gaussians to the current timestamp. The semantic features are $L_2$-normalized
\begin{equation}
\hat f_i=\frac{f_i}{\|f_i\|_2+\epsilon},
\end{equation}
and splatted using the differentiable Gaussian renderer to produce a dense semantic feature map
\[
F\in\mathbb{R}^{32\times H\times W}.
\]

Semantic supervision is obtained from 2D segmentation masks (e.g., produced by SAM). At each iteration we sample $N=10{,}000$ pixels and $M=50$ masks. Let $C$ denote the pixel–mask correspondence matrix and $C_F$ the cosine similarity matrix of rendered features. We optimize the features using a soft contrastive objective
\begin{equation}
\mathcal{L}_{semantic}
=
\mathcal{L}_{pos}(C,C_F)
+
\mathcal{L}_{neg}(C,C_F)
+
\lambda_{rfn}(1-\|F\|_2)^2 ,
\end{equation}
where the final term prevents feature collapse ($\lambda_{rfn}=1.0$).

Because semantic learning operates only on the compact persistent subset $\mathcal{G}_p$, the number of optimized primitives is reduced by approximately $48\times$ compared to monolithic 4DGS representations. This preserves temporal identity while enabling nearly $10\times$ faster semantic training and inference.

\subsection{4D Semantic Clustering}

Once semantic features are optimized, we extract discrete object instances through feature-space clustering. Let$
F_p\in\mathbb{R}^{N\times32}
$
denote the normalized semantic features of the persistent Gaussians. To reduce computational cost, we randomly sample $2\%$ of these features
\[
F_{sample}\in\mathbb{R}^{N'\times32}, \quad N'=0.02N .
\]

Density-based clustering (DBSCAN)~\cite{ester1996density} is applied to the sampled features to identify semantic groups. Cluster centroids are then computed and normalized to form
\[
C\in\mathbb{R}^{K\times32}.
\]

All persistent Gaussians are assigned to clusters using cosine similarity:
\begin{equation}
S = F_p C^\top,
\qquad
\text{ID}_i=\arg\max_k S_{i,k}.
\end{equation}

Because cluster labels are attached to persistent Gaussians and propagated through the deformation field $\Phi_g$, object identities remain temporally consistent. Objects can therefore be rendered at arbitrary timestamps by isolating the corresponding primitives.

\subsection{Open-Vocabulary 4D Segmentation}

Finally, we support open-vocabulary 4D segmentation from a single text prompt. Given a prompt (e.g., ``person''), we render the first frame and apply Grounding DINO~\cite{liu2023grounding} to obtain bounding boxes, which are refined with SAM~\cite{kirillov2023segment} to produce a binary mask $M_{text}$.

Masked pixels are unprojected into 3D using the rendered depth map:
\begin{equation}
P=[u,v,z,1]^T\mathbf{K}^{-1}\mathbf{E}^{-1}.
\end{equation}

The persistent Gaussians are deformed to the same timestamp and matched to the unprojected points via nearest-neighbor search. The corresponding semantic cluster is identified through majority voting.

To refine object boundaries, we compute the mean feature of the cluster
\begin{equation}
\bar f_{target}=
\frac{1}{|\mathcal{C}_{target}|}
\sum_{i\in\mathcal{C}_{target}}\hat f_i ,
\end{equation}
and retain only Gaussians whose cosine similarity exceeds $\tau_{score}=0.95$:
\begin{equation}
\mathcal{C}_{final}=
\{i\in\mathcal{C}_{target}\mid
\hat f_i\cdot\bar f_{target}\ge\tau_{score}\}.
\end{equation}

The selected primitives can then be rendered across all timestamps using the deformation field $\Phi_g$, producing temporally consistent object masks and RGB renderings without per-frame tracking or optical flow.
\section{Additional Experimental Analysis}

\subsection{Compactness and Persistent Motion}
\begin{figure}[h]
  \centering
  \includegraphics[width=0.8\linewidth]{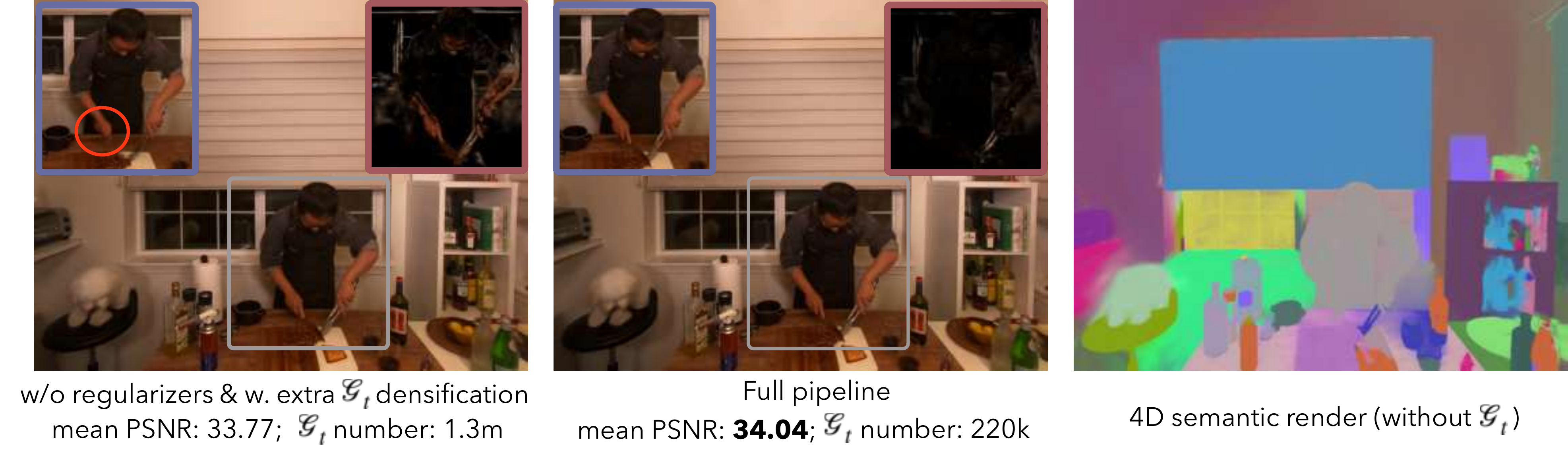}
  \caption{Over-parameterized transients degrade persistent motion.}
  \label{fig:compactness_supp}
\end{figure}

Multi4D targets accurate static--motion--transient decomposition through cross-branch self-regularization. Compactness is not only a storage objective: it is part of the decomposition quality. As shown in~\cref{fig:compactness_supp}, excessive transient parameterization disrupts the decomposition, increases storage, lowers PSNR, and degrades persistent motion. Unlike render-first pipelines that reuse one primitive set for both photometric fidelity and motion correspondence, Multi4D optimizes downstream semantics only on the stable persistent subset, preventing transient-induced inconsistency.

\subsection{Sensitivity Analysis}
Multi4D achieves compact subset specialization through appearance competition, branch-specific densification/pruning, and utility-based lifting, rather than fixed motion-score assignment. Pruning, regrowth, and cross-branch lifting progressively correct wrong allocations, enabling one fixed parameter set for all scenes. As shown in~\cref{tab:sensitivity_analysis_supp}, sweeps over regularizers, lifting count, and mask-score threshold remain close to the reference setting, indicating robustness to exact hyperparameters. The lifting threshold is a coarse indicator of activated foreground Gaussians, not a transient classifier; final allocation is determined by shared competition, densification, and pruning.

\begin{table}[h]
    \centering
    \caption{Sensitivity analysis on Neu3D. We report $\Delta$ mean peak PSNR relative to the reference setting.}
    \label{tab:sensitivity_analysis_supp}
    \resizebox{\linewidth}{!}{%
    \begin{tabular}{lcccc|cccc|ccccc|cccccc}
    \toprule
    & \multicolumn{4}{c|}{$\mathcal{L}_\alpha$ (alpha regularizer)}
    & \multicolumn{4}{c|}{$\mathcal{L}_{\mathrm{TV}}$ (depth regularizer)}
    & \multicolumn{5}{c|}{$K$ samples in lifting}
    & \multicolumn{6}{c}{Lifting mask threshold $\tau$} \\
    \cmidrule(lr){2-5}
    \cmidrule(lr){6-9}
    \cmidrule(lr){10-14}
    \cmidrule(lr){15-20}
    Setting
    & 0 & 0.005 & \textbf{0.01} & 0.02
    & 0 & 0.005 & \textbf{0.01} & 0.02
    & 500 & 1000 & \textbf{2000} & 4000 & 10000
    & 0 & 0.02 & \textbf{0.05} & 0.07 & 0.10 & 0.50 \\
    \midrule
    $\Delta$ PSNR
    & -0.08 & -0.05 & 0.00 & -0.17
    & -0.16 & -0.10 & 0.00 & -0.03
    & -0.16 & -0.08 & 0.00 & -0.03 & -0.22
    & -0.14 & -0.06 & 0.00 & -0.01 & -0.03 & -0.28 \\
    \bottomrule
    \end{tabular}%
    }
\end{table}

\subsection{LPIPS Evaluation}
We additionally report LPIPS to quantify perceptual rendering quality on both multi-view and monocular benchmarks. For the multi-view Technicolor and Neu3D datasets, we report LPIPS-Alex, while for the monocular NeRF-DS dataset, we report LPIPS-VGG following the corresponding evaluation protocol. As shown in~\cref{tab:lpips_all_supp}, Multi4D achieves strong perceptual quality across Technicolor, Neu3D, and NeRF-DS.

\begin{table}[h]
  \centering
  \caption{Per-scene LPIPS on multi-view and monocular datasets.}
  \label{tab:lpips_all_supp}
  \scriptsize
  \setlength{\tabcolsep}{2.3pt}
  \resizebox{0.98\linewidth}{!}{%
  \begin{tabular}{l ccccc c | cccccc c}
  \toprule
  & \multicolumn{6}{c|}{Technicolor LPIPS $\downarrow$}
  & \multicolumn{7}{c}{Neu3D LPIPS $\downarrow$} \\
  \cmidrule(lr){2-7}
  \cmidrule(lr){8-14}
  Method
  & Birthday & Fabien & Painter & Theater & Train & Mean
  & \makecell{Cut\\Beef}
  & \makecell{Cook\\Spinach}
  & \makecell{Sear\\Steak}
  & \makecell{Flame\\Steak}
  & \makecell{Flame\\Salmon}
  & \makecell{Coffee\\Martini}
  & Mean \\
  \midrule
  DyNeRF & 0.0668 & 0.2417 & 0.1464 & 0.1881 & \third{0.0670} & 0.1400 & - & - & - & - & - & - & - \\
  NeRFPlayer$^2$ & - & - & - & - & - & - & 0.1440 & 0.1130 & 0.1380 & 0.0880 & 0.0980 & 0.0850 & 0.1110 \\
  HyperReel & 0.0531 & 0.1864 & 0.1173 & \second{0.1154} & 0.0723 & 0.1090 & 0.0840 & 0.0890 & 0.0770 & 0.0780 & 0.1360 & 0.1270 & 0.0985 \\
  HexPlane$^{1,2}$ & - & - & - & - & - & - & 0.0800 & 0.0820 & 0.0700 & 0.0660 & 0.0780 & - & 0.0750 \\
  \midrule
  Def-3DGS & 0.0775 & 0.1851 & 0.1302 & 0.1795 & 0.2040 & 0.1553 & 0.0551 & 0.0519 & 0.0416 & 0.0418 & 0.0804 & 0.0855 & 0.0594 \\
  4DGaussian & 0.0846 & 0.1868 & 0.1500 & 0.1825 & 0.2194 & 0.1647 & 0.0538 & 0.0522 & 0.0415 & 0.0402 & 0.0818 & 0.0830 & 0.0588 \\
  DeGauss & - & - & - & - & - & - & \third{0.0423} & 0.0413 & 0.0359 & 0.0347 & 0.0684 & \second{0.0625} & 0.0475 \\
  E-D3DGS & \third{0.0506} & 0.1689 & 0.0903 & 0.1493 & 0.0976 & \third{0.1114} & \first{0.0336} & \first{0.0338} & \third{0.0301} & \second{0.0284} & \first{0.0535} & \first{0.0417} & \first{0.0369} \\
  \midrule
  4DGS & 0.0629 & \third{0.1555} & \third{0.1125} & 0.1653 & 0.0985 & 0.1189 & 0.0470 & 0.0489 & 0.0411 & 0.0389 & 0.0832 & 0.0847 & 0.0573 \\
  STG$^2$ & \second{0.0413} & \second{0.1140} & \second{0.0963} & \third{0.1332} & \second{0.0380} & \second{0.0846} & \second{0.0367} & \third{0.0374} & \second{0.0295} & \third{0.0290} & \second{0.0630} & \third{0.0692} & \third{0.0441} \\
  \midrule
  \textbf{Ours} & \first{0.0403} & \first{0.1100} & \first{0.0868} & \first{0.0824} & \first{0.0327} & \first{0.0704} & \second{0.0367} & \second{0.0373} & \first{0.0294} & \first{0.0282} & \second{0.0620} & 0.0706 & \second{0.0440} \\
  \bottomrule
  \end{tabular}}

  \vspace{0.5em}
  \setlength{\tabcolsep}{3pt}
  \resizebox{0.8\linewidth}{!}{%
  \begin{tabular}{lccccccc|c}
  \toprule
  \multicolumn{9}{c}{Monocular NeRF-DS LPIPS $\downarrow$} \\
  \midrule
  Method & As & Basin & Bell & Cup & Plate & Press & Sieve & Mean \\
  \midrule
  NeRF-DS & 0.2150 & 0.2508 & 0.2921 & \third{0.1707} & \third{0.2974} & 0.2552 & 0.2001 & 0.2402 \\
  HyperNeRF & 0.2390 & 0.2497 & 0.2999 & 0.2302 & 0.3166 & 0.2810 & 0.2142 & 0.2615 \\
  \midrule
  Def-3DGS & \second{0.1850} & \first{0.1924} & \third{0.2767} & \second{0.1658} & 0.3599 & \first{0.1964} & \first{0.1643} & \third{0.2201} \\
  4DGaussian & \third{0.2038} & \third{0.2196} & \second{0.2061} & 0.1792 & \second{0.2721} & \third{0.2255} & \third{0.1745} & \second{0.2115} \\
  \midrule
  4DGS & 0.2671 & 0.3312 & 0.2932 & 0.3429 & 0.4374 & 0.3696 & 0.3318 & 0.3390 \\
  STG & 0.3196 & 0.3189 & 0.2755 & 0.2910 & 0.4097 & 0.3154 & 0.2715 & 0.3145 \\
  \midrule
  \textbf{Ours} & \first{0.1827} & \second{0.2054} & \first{0.1682} & \first{0.1627} & \first{0.2288} & \second{0.2140} & \second{0.1704} & \first{0.1903} \\
  \bottomrule
  \end{tabular}}
\end{table}

\subsection{Initialization Robustness}
\begin{figure}[h]
  \centering
  \includegraphics[width=1\linewidth]{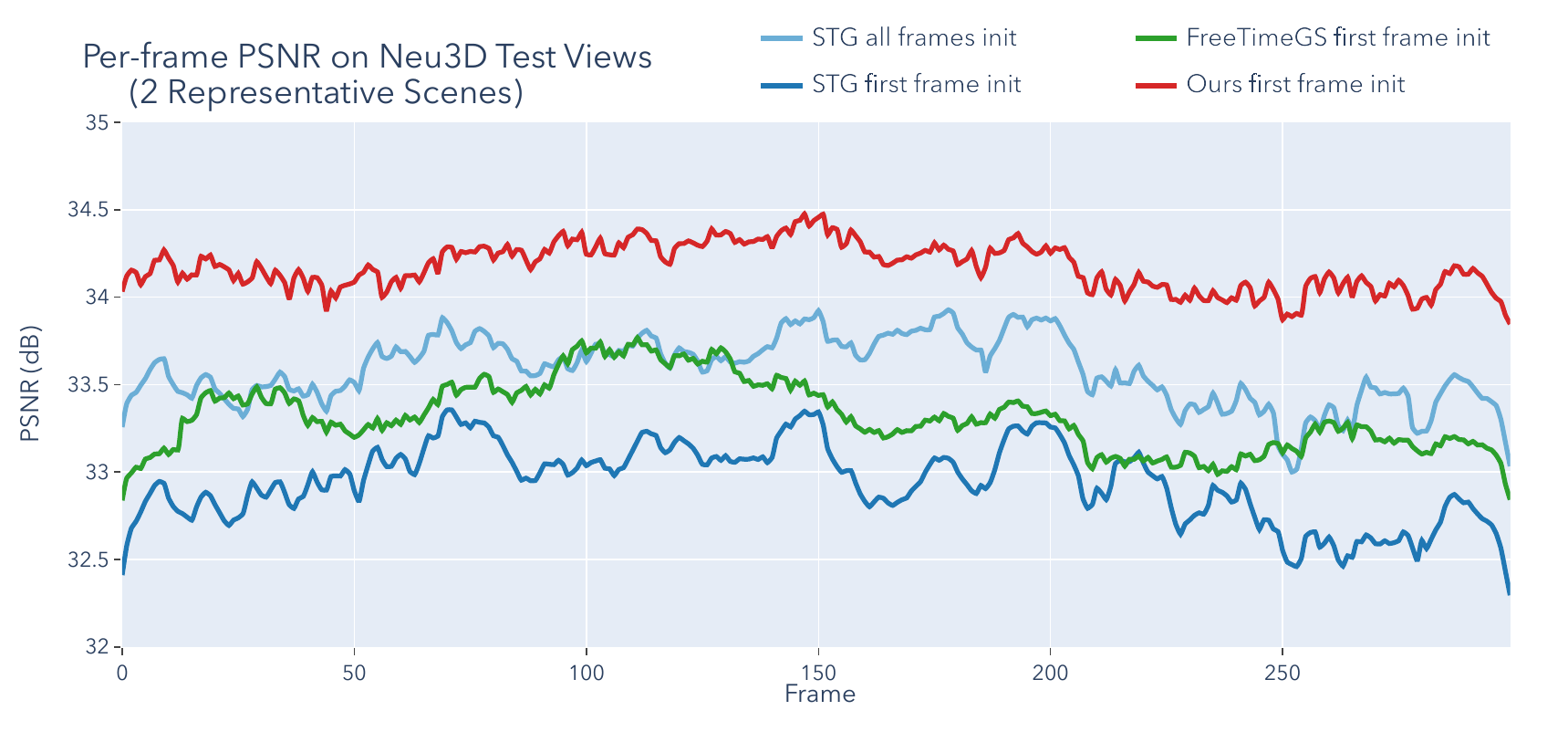}
  \caption{Per-frame PSNR on Neu3D test views. Mean PSNR: STG first-frame init 32.94, STG all-frame init 33.58, FreeTimeGS first-frame init 32.95 (our reproduction; official code unavailable), and Multi4D first-frame init 34.17.}
  \label{fig:init_supp}
\end{figure}

\cref{fig:init_supp} compares per-frame PSNR on Neu3D test views. With only first-frame initialization, Multi4D achieves higher and more stable PSNR than STG with first/all-frame initialization and our FreeTimeGS reproduction (official code unavailable). FreeTimeGS reports dense 4D position/velocity initialization, whereas Multi4D uses only first-frame initialization; this indicates that the gain comes from the three-branch representation and persistent motion backbone rather than dense temporal initialization.

\section{Extended Discussion}

\subsection{Additional Discussion on Related Work}

\textbf{NeRF-based decomposition.}
D$^2$-NeRF~\cite{wu2022d} and NeRFPlayer~\cite{song2023nerfplayer} also explore scene decomposition for dynamic reconstruction, but rely on NeRF-based implicit representations. By contrast, Gaussian-based methods such as DeGauss~\cite{wang2025degauss} and our Multi4D achieve superior training and rendering efficiency, and naturally benefit from explicit densification for compact, self-regularized scene modeling. In addition, the explicit Gaussian representation enables direct cross-subset transfer and lifting, allowing Multi4D to incorporate motion-aware primitives more effectively for faster modeling with better geometry-motion consistency.

\textbf{Gaussian decomposition and layering.}
Prior Gaussian decomposition methods such as Swift4D~\cite{wu2025swift4d}, MAPo~\cite{jiao2026mapo}, and motion-layering approaches~\cite{dai20254d,liu2026dynamics} only work on static-camera multiview datasets, and
  often start from external cues such as pixel differences, optical flow, or SAM masks. These cues are effective in static camera multi-view settings, but often fail for
  monocular/moving-camera capture or non-motion appearance changes such as fire and reflections. Multi4D instead treats decomposition as an intrinsic outcome of Gaussian
  optimization: static, persistent dynamic, and transient branches compete in a shared renderer, while each branch performs its own densification and pruning, so the
  decomposition is continuously revised through branch-specific growth/removal rather than fixed by an external motion score, improving robustness to floaters. This also addresses the expressiveness-regularization trade-off at representation level: prior work increases deformation capacity for multiview fidelity, but this can weaken regularization
  needed for stable monocular/sparse-view reconstruction. Multi4D decouples persistent motion from transient residuals, keeping coherent motion regularized while
  capturing short-lived details compactly.
This allows the same design to work well in both multiview
  and monocular settings.
\subsection{Additional Qualitative Analysis and Limitations}
\begin{figure}[h]
  \centering
  \includegraphics[width=0.85\linewidth]{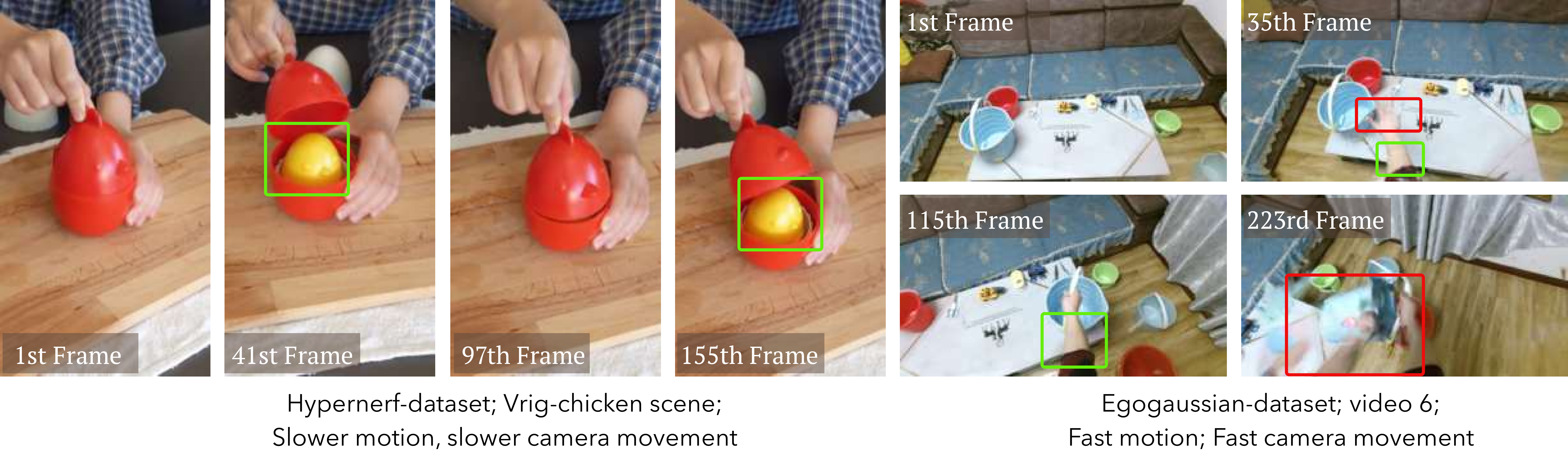}
  \caption{New-object success/failure cases on monocular datasets~\cite{park2021hypernerf,zhang2024egogaussian}.}
  \label{fig:newobject_supp}
\end{figure}

\cref{fig:newobject_supp} discusses new-object cases on monocular datasets. Newly appearing content is first captured by the transient branch and can become persistent if repeatedly observed. Very fast camera/object motion remains challenging. Low-texture regions, such as the Painter sequence, provide weak appearance-only separation cues despite high PSNR, and boundary confusions mainly reflect the smoothness bias of persistent deformation.

\subsection{All-Point Tracking and Feed-Forward Dynamic Scene Representation}

Multi4D identifies a key limitation of existing dynamic Gaussian representations: high-fidelity rendering and temporally consistent correspondence are often treated as competing objectives. Our hybrid representation addresses this by disentangling static structure from dynamic content, and persistent motion from transient appearance variations. Through bottom-up, self-regularized optimization, Multi4D yields a compact representation that preserves correspondence-aware persistent motion while retaining high-quality scene reconstruction.

This perspective is also relevant to feed-forward dynamic Gaussian reconstruction. Most existing feed-forward dynamic Gaussian representations rely on per-frame 3DGS~\cite{liang2024feed} or dense 4DGS~\cite{luo2026instant4d}, which can fall into the over-parameterization regime and provide limited explicit motion correspondence. In particular, transient effects and disocclusions make it difficult to maintain stable point identities over time. Multi4D suggests that a hybrid decomposition into persistent and transient components could provide a useful structural prior for feed-forward models.

Extending Multi4D to a feed-forward setting, and evaluating point/Gaussian tracking accuracy on diverse egocentric and dynamic-scene benchmarks~\cite{pan2023aria,wang2023pov,zheng2023point}, is therefore an interesting direction for future work.

\end{document}